\documentclass[10pt,compsoc,journal]{IEEEtran}
\usepackage{array}
\usepackage[caption=false,font=normalsize,labelfont=sf,textfont=sf]{subfig}
\usepackage{stfloats}
\usepackage{verbatim}
\usepackage{cite}

\usepackage{hyperref}
\usepackage{url}
\usepackage{algorithm}
\usepackage{algorithmic}
\usepackage{graphicx}%
\usepackage{multirow}%
\usepackage{amsmath,amssymb,amsfonts}%
\usepackage{amsthm}%
\usepackage{mathrsfs}%
\usepackage{xcolor}%
\usepackage{textcomp}%
\usepackage{manyfoot}%
\usepackage{booktabs}%
\usepackage{listings}%
\usepackage{anyfontsize}
\usepackage{color}
\usepackage{bbding}
\usepackage{longtable}
\usepackage{supertabular}
\usepackage{colortbl}
\usepackage{wrapfig} 
\usepackage{mathrsfs}
\usepackage{pifont}

\newtheorem{definition}{Definition}
\newtheorem{observation}{Observation}

\begin{document}

\title{Task-Specific Directions: Definition, Exploration, and Utilization in Parameter Efficient Fine-Tuning
}

\author{
Chongjie Si, Zhiyi Shi, Shifan Zhang, Xiaokang Yang,~\IEEEmembership{~Fellow,~IEEE}, Hanspeter Pfister,~\IEEEmembership{~Fellow,~IEEE}, Wei Shen
\thanks{Corresponding author: Wei Shen.}
\thanks{C. Si, X. Yang, and W. Shen are with
MoE Key Lab of Artificial Intelligence, AI Institute, Shanghai Jiao Tong University, Shanghai, China.}
\thanks{S. Zhang is with the School of Electronic Information and Electrical Engineering, Shanghai Jiao Tong University, Shanghai, China.}
\thanks{Z. Shi and H. Pfister are with the School of Engineering and Applied Sciences, Harvard University, Cambridge, MA 02138, USA.}
\thanks{Email: \{chongjiesi, zhangshifan, xkyang, wei.shen\}@sjtu.edu.cn, \{zhiyis, pfister\}@g.harvard.edu }
}

\markboth{A SUBMISSION TO IEEE TRANSACTION ON PATTERN ANALYSIS AND MACHINE INTELLIGENCE}
{Shell \MakeLowercase{\textit{et al.}}: A Sample Article Using IEEEtran.cls for IEEE Journals}


\IEEEtitleabstractindextext{
\begin{abstract}
Large language models demonstrate impressive performance on downstream tasks, yet requiring extensive resource consumption when fully fine-tuning all parameters. 
To mitigate this, Parameter Efficient Fine-Tuning (PEFT) strategies, such as LoRA, have been developed. 
In this paper, we delve into the concept of task-specific directions (TSDs)—critical for transitioning large models from pretrained states to task-specific enhancements in PEFT. 
We propose a framework to clearly define these directions and explore their properties, and practical utilization challenges.
We then introduce a novel approach, LoRA-Dash, which aims to maximize the impact of TSDs during the fine-tuning process, thereby enhancing model performance on targeted tasks.
Additionally, based on our exploration of TSD, we focus on an important issue in PEFT: the initialization of LoRA. 
While some works have pointed out the significance of initialization for LoRA's performance and proposed various strategies, these methods are often empirical and not task-specific.
To address this issue, we propose LoRA-Init. Starting from TSD, we identify the directions that require the most adjustment during fine-tuning for downstream tasks. 
By initializing the matrices in LoRA with these directions, LoRA-Init significantly enhances LoRA’s performance.
Moreover, we can combine LoRA-Dash and LoRA-Init to create the final version of LoRA based on TSDs, which we refer to as LoRA-TSD.
Extensive experiments have conclusively demonstrated the effectiveness of these methods, and in-depth analyses further reveal the underlying mechanisms of these methods. 
The codes are available at \url{https://github.com/Chongjie-Si/Subspace-Tuning}.
\end{abstract}

}
\maketitle
\IEEEdisplaynontitleabstractindextext
\IEEEpeerreviewmaketitle

\section{Introduction}
Large language models (LLMs) \cite{liu2019roberta, qin2023chatgpt,he2021debertav3,touvron2023llama,devlin2018bert,touvron2023llama2,llama3modelcard} have exhibited superior performance in a variety of natural language processing (NLP) tasks, including commonsense reasoning \cite{sap2020commonsense} and natural language understanding \cite{wang2018glue}. Despite the advantages, the practice of fully fine-tuning these models, which often comprise hundreds of millions to hundreds of billions of parameters, requires substantial computational resources and incurs significant memory costs \cite{touvron2023llama,ma2024segment,raffel2020exploring}. This extensive resource requirement restricts the practical deployment of LLMs across diverse scenarios.

To address this issue, parameter efficient fine-tuning (PEFT) \cite{zhang2022adaptive,hu2021lora,si2024flora,houlsby2019parameter,zaken2021bitfit,si2024see} has been a focal point of recent advancements in adapting LLMs with minimal computational and GPU memory costs. It aims to optimize the number of adjustable parameters to improve task performance without changing the original structure of the model. 
Among various PEFT methods, LoRA (low-rank adaptation) \cite{hu2021lora} stands out. Since the effectiveness of fine-tuning can be ascribed to the reparameterization of a ``low-dimensional manifold'' \cite{aghajanyan2020intrinsic,li2018measuring}, it suggests that the changes to linear model weights can be characterized by a low-rank structure. 
Specifically, for each weight matrix $\mathbf{W}\in\mathbb{R}^{n\times m}$, LoRA models its changes $\Delta\mathbf{W}\in\mathbb{R}^{n\times m}$ using two low-rank matrices, $\mathbf{A}\in\mathbb{R}^{n\times r}$ and $\mathbf{B}\in\mathbb{R}^{r\times m}$, with $r\ll \{n,m\}$ to achieve parameter efficiency. The change $\Delta\mathbf{W}$ is added to the pre-trained weights $\mathbf{W}$, and only $\mathbf{A}$ and $\mathbf{B}$ are updated during training. Due to LoRA’s flexibility, its applications have become widespread \cite{rombach2022highstablediffusion, ma2024segment}.

LoRA underscores that $\Delta\mathbf{W}$ enhances directions in $\mathbf{W}$ that, although not pivotal for pretraining, are indispensable for specific downstream tasks, referring to these as ``task-specific directions''.
Indeed, we suggest that these directions serve as intuitive representations of the low-dimensional manifold, which indicates the significance of task-specific directions for the success of fine-tuning.
However, despite that LoRA refers to or even points out the significance of these directions, no existing studies have concretely defined or systematically pinpointed them, let alone outlined strategies for their effective utilization. 
This not only suppresses LoRA’s potential but also hinders a deeper understanding of the fundamental mechanisms underlying fine-tuning.

The chief goal of this paper is to bridge this gap and further unleash the potential of task-specific directions in fine-tuning, we accomplished two key advancements, which constitute the contributions of this paper for more details on the structure and contributions of this paper. 
First, we propose a framework to provide a precise definition of task-specific directions, exploring the properties and practical utilization challenges of these directions.
Second, building on the comprehensive analysis, we introduce a novel method, LoRA-Dash, which identifies task-specific directions during training and proactively utilizes their influence for different usages. 
Extensive experiments demonstrate the effectiveness of LoRA-Dash, and detailed analyses provide us a more comprehensive understanding of LoRA-Dash.

This paper is an extension of our preliminary work \cite{si2025unleashing}.
In our previous work, we primarily focused on exploring TSD and its simple application, LoRA-Dash. 
In this work, we extend this exploration to address a more practical problem: the initialization of LoRA. 
Many studies \cite{meng2024pissa,hameed2024rosa,wang2024milora} have highlighted the critical role of initialization in the performance of LoRA.
As a result, various empirical or experimental initialization strategies \cite{meng2024pissa, wang2024loraga} have been proposed, such as utilizing the top singular components of the weight matrix $\mathbf{W}$ or leveraging gradient information. 
However, these initialization methods are often not task-specific.
Based on the properties of TSDs, in this work, we propose a novel LoRA initialization method, LoRA-Init.
We first identify the directions in which the model requires the most adjustment when fine-tuning to downstream tasks, and by initializing LoRA with TSDs, we can achieve faster convergence and better performance. 
Furthermore, by integrating LoRA-Dash and LoRA-Init, we can develop the ultimate version of LoRA based on TSDs, which we designate as LoRA-TSD.
LoRA-TSD utilizes TSD to initialize LoRA, training the directions that require the most modification while fixing those that need less adjustment. 
Meanwhile, LoRA-Dash directly simulates these TSDs, fully harnessing the potential of the TSDs.
Through extensive experiments, we demonstrate the effectiveness of these methods.

Our main contributions are summarized as follows:
\begin{itemize}
    \item We started from scratch, gradually building a comprehensive framework to rigorously define TSDs. Based on this framework, we delved into the various properties of TSDs and identified the challenges associated with utilizing them, along with the corresponding strategies to overcome these difficulties.
    \item We propose a new plugin based on the properties of TSDs, LoRA-Dash. It aims to identify TSDs and directly simulate their corresponding changes, thereby enhancing the performance of methods like LoRA on downstream tasks. It demonstrates strong scalability and performance across a variety of methods.
    \item We propose LoRA-Init to use TSDs to address the practical issue of initializing LoRA. By identifying TSDs, we directly initialize the directions most critical for downstream tasks into the LoRA matrices, resulting in improved performance.
    \item We propose LoRA-TSD, the final version that fully harnesses the power of TSD, integrating both TSD-based initialization and direct simulation of TSD directions to optimize performance.
\end{itemize}
The rest of our paper is organized as follows.
We first review the description and exploration of TSDs within the LoRA framework in Sec. \ref{sec: lora contradiction}. 
In Sec. \ref{sec: a framework}, we then propose a novel framework that clearly defines what TSDs are and further explores their properties. Additionally, we highlight the challenges of applying TSDs.
To further unleash the potential of TSDs, we propose three new methods, LoRA-Dash, LoRA-Init for LoRA initialization, which leverages TSDs to enhance performance on downstream tasks, and LoRA-TSD, in Sec. \ref{sec: lora-dash}. 
We conduct various experiments to further analyze the underlying mechanisms of these methods in Secs. \ref{sec: performance}-\ref{sec: understanding lora-dash}, and finally give the conclusion in Sec. \ref{sec conclusion}.


\section{Background} \label{sec: lora contradiction}

\subsection{Parameter Efficient Fine-tuning}
LLMs' huge complexity and computational demands with billions of parameters create significant challenges for adapting them to specific downstream tasks \cite{xu2023parameter,han2024parameter}. Parameter Efficient Fine-Tuning (PEFT) offers an effective solution by reducing the parameters and memory needed for adaptation to a variety of downstream tasks while maintaining performance levels similar to full fine-tuning \cite{si2024see,ding2023parameter}. Current PEFT methods can be roughly divided into three categories \cite{si2024see,liu2024dora}: adapter-based \cite{houlsby2019parameter,chen2022adaptformer,luo2023towards,he2021towards,mahabadi2021parameter,karimi2021compacter}, prompt-based \cite{lester2021power,razdaibiedina2023residual,wang2023non,shi2023dept,fischer2024prompt}, and low-rank matrix decomposition-based \cite{hu2021lora,liu2024dora,hyeon2021fedpara,qiu2023controlling,renduchintala2023tied,kopiczko2023vera,yeh2023navigating,zhang2022adaptive}. 
The first category of methods enhances performance by integrating linear modules with existing layers, either sequentially or concurrently. 
The second category focuses on refining trainable vectors by adding soft tokens (prompts) to the initial input.
The third type, introduced by LoRA \cite{hu2021lora}, employs low-rank adaptation to model weight changes during fine-tuning and can be combined with pre-trained weights.

\subsection{LoRA: Low-rank Adaptation}

Based on findings that updates to the weights typically exhibit a low intrinsic rank \cite{aghajanyan2020intrinsic,li2018measuring}, LoRA models the changes $\Delta\mathbf{W}\in\mathbb{R}^{n\times m}$ for each layer's weights $\mathbf{W}\in\mathbb{R}^{n\times m}$ as $\Delta\mathbf{W} = \mathbf{A}\mathbf{B}$, where $\mathbf{A}\in\mathbb{R}^{n\times r}$, $\mathbf{B}\in\mathbb{R}^{r\times m}$ with the rank $r \ll \{n,m\}$ to achieve parameter efficiency. For the original output $\mathbf{h}=\mathbf{W}\mathbf{x}$, the modified forward pass is
\begin{equation}
    \mathbf{h}=\mathbf{W}\mathbf{x} + \Delta\mathbf{W}\mathbf{x} = (\mathbf{W}+\mathbf{A}\mathbf{B})\mathbf{x}.
\end{equation}
In the training initialization for LoRA, matrix $\mathbf{A}$ is commonly set with Kaiming distribution \cite{he2015delving}, and matrix  $\mathbf{B}$ is initialized with zeros, which sets the initial $\Delta\mathbf{W}$ to zero at beginning. During training, LoRA only updates the low-rank matrices $\mathbf{A}$ and $\mathbf{B}$ with $\mathbf{W}$ being frozen.
During inference, the low-rank matrices are integrated into the $\mathbf{W}$, resulting in no additional costs.

\subsection{How Do Task-specific Directions Arise in LoRA?}\label{sec: supp tsd arise}

To explore the relationship between the learned $\Delta\mathbf{W}$ and the original weights $\mathbf{W}$, LoRA initially applies SVD to $\Delta\mathbf{W}$ to extract its left and right singular vectors, $\mathbf{U}$ and $\mathbf{V}$. 
\textbf{\textit{LoRA first validates that the directions corresponding to the top singular vectors of $\Delta\mathbf{W}$ tend to overlap significantly across different ranks}} (conclusion 1).
Subsequently, LoRA projects $\mathbf{W}$ onto the $r$-dimensional subspace defined by $\Delta\mathbf{W}$, calculating $\mathbf{U}^\mathsf{T} \mathbf{W}\mathbf{V}$ and its Frobenius norm $\|\mathbf{U}^\mathsf{T} \mathbf{W}\mathbf{V}\|_F$.
This norm is also computed by replacing $\mathbf{U}$ and $\mathbf{V}$ with those from $\mathbf{W}$ or a random matrix for comparison. 
The results reveal that $\|\mathbf{U}^\mathsf{T} \mathbf{W}\mathbf{V}\|_F$ when $\mathbf{U}$ and $\mathbf{V}$ are top singular vectors derived from $\Delta\mathbf{W}$ or $\mathbf{W}$ is much greater than when they are derived from a random matrix, suggesting a stronger correlation of $\Delta\mathbf{W}$ with $\mathbf{W}$ than with a random matrix. 
\textbf{\textit{This indicates that the features amplified by $\Delta\mathbf{W}$ already present in $\mathbf{W}$}}.

Additionally, LoRA introduces the ``feature amplification factor'' to measure the extent of feature enhancement, defined as ${\|\Delta\mathbf{W}\|_F} / {\|\mathbf{U}^\mathsf{T} \mathbf{W}\mathbf{V}\|_F}$. 
The factor is significantly higher when $\mathbf{U}$ and $\mathbf{V}$ are top singular vectors derived from $\Delta\mathbf{W}$ compared to when they are those derived from $\mathbf{W}$, \textbf{\textit{suggesting that $\Delta\mathbf{W}$ only boosts directions that are not emphasized in $\mathbf{W}$}}. 
Moreover, a larger $r$  ($r=64$) yields a much lower amplification factor than a smaller $r$ ($r=4$), \textbf{\textit{implying that the number of ``task-specific directions'' is small}}. 
Through extensive experiments, LoRA determines that $\Delta\mathbf{W}$ potentially amplifies important directions for specific downstream tasks that were learned but not emphasized during general pre-training, and refers to them as ``task-specific'' directions (TSDs, plural).

\subsection{Rethinking ``Task-Specific Directions'' in LoRA}\label{sec: pre rethinking}

LoRA has drawn three conclusions related to TSDs:

\begin{enumerate}
    \item TSDs are the directions of top singular vectors derived from the learned $\Delta\mathbf{W}=\mathbf{A}\mathbf{B}$. TSDs consistently exhibit significant overlap across different $r$ settings used in the learning of  $\Delta\mathbf{W}$, where $r$ is the rank configuration of $\Delta\mathbf{W}$.
    \item TSDs are not the directions of top singular vectors derived from the pretrained weights $\mathbf{W}$.
    \item TSDs are certain directions that have already been learned by $\mathbf{W}$ but were not emphasized.
\end{enumerate}

However, even the first conclusion of LoRA presents significant contradictions. For a specific task, TSDs should indeed remain consistent and not depend on learned $\Delta\mathbf{W}$. Moreover, revisiting the first and the third conclusions, we could deduce that the top singular directions of $\Delta\mathbf{W}$ are among the directions of $\mathbf{W}$. Nonetheless, in practice, the singular directions of $\Delta\mathbf{W}$ are unlikely to align exactly with any specific directions in $\mathbf{W}$\footnote{For more details on what they did and why they drew the conclusions, we encourage readers to refer to their original paper \cite{hu2021lora}, specifically Section 7.}.

Indeed, we suggest that ``TSDs'' are intuitive depictions of the low-dimensional manifold \cite{aghajanyan2020intrinsic,li2018measuring}.
Given that the most impactful transformations occur within the low-dimensional manifold during fine-tuning, TSDs are also significant in PEFT.
However, the description of TSDs in LoRA presents several contradictions that contribute to confusion.
This confusion complicates the identification of TSDs and the subsequent utilization of them.

We believe that these issues are mainly due to an unclear definition of TSDs. 
Consequently, to clearly define TSDs and enhance their practical application, we have discarded all previous definitions utilized by LoRA. 
We aim to undertake a comprehensive reconstruction of the entire framework from scratch, ensuring that TSDs are not only well-defined but also effectively utilized in further applications.

\section{Build a Framework: Understanding Task-specific Directions} \label{sec: a framework}

\subsection{Preliminaries}
Considering a matrix $\mathbf{A}\in \mathbb{R}^{n \times m}$ where $n<m$, it can be decomposed using SVD as $\mathbf{A}=\mathbf{U}\mathbf{\Sigma}\mathbf{V}^\mathsf{T}$. Here, $\mathbf{\Sigma}={\rm diag}(\sigma_1,\dots,\sigma_n)$ is the diagonal matrix of singular values, and $\mathbf{U}$ and $\mathbf{V}$ are the left and right singular vectors, respectively. 
This decomposition can be further expressed as 
\begin{equation}
    \mathbf{A} = \sum_{i=1}^n \sigma_i \mathbf{u}_i \mathbf{v}_i^\mathsf{T}.
\end{equation}
Since $\{\mathbf{u}_i|i=1,2\dots,n\}$ forms the orthogonal bases for $\mathbb{R}^n$ and $\{\mathbf{v}_i|i=1,2\dots,n\}$ for $\mathbb{R}^m$, the matrix space $\mathbb{R}^{n \times m}$ can be spanned by the bases $\{\mathbf{u}_i\mathbf{v}_j^\mathsf{T}|i=1,2\dots,n, j=1,2\dots,m\}$, which are also orthogonal. 
Thus, matrix $\mathbf{A}$ can be seen as a matrix in a subspace of the original $\mathbb{R}^{n \times m}$ space, spanned by a set of linearly independent bases $\{\mathbf{u}_i\mathbf{v}_i^\mathsf{T}|i=1,2\dots,n\}$.
\begin{definition}
    For a matrix $\mathbf{A}\in\mathbb{R}^{n\times m}$ ($n<m$) with its left and right singular vectors represented by matrices $\mathbf{U}$ and $\mathbf{V}$, respectively, the bases of $\mathbf{A}$ are defined as follows:
    \begin{itemize}
        \item \textbf{Core Bases}: The core bases of the matrix $\mathbf{A}$ are defined as $\{\mathbf{u}_i \mathbf{v}_i^\mathsf{T}|i=1,2\dots,n\}$, where each $\mathbf{u}_i \mathbf{v}_i^\mathsf{T}$ is a rank-one matrix formed by the outer product of singular vectors $\mathbf{u}_i$ and $\mathbf{v}_i$.
        \item \textbf{Global Bases}: The global bases of the matrix $\mathbf{A}$ are defined as $\{\mathbf{u}_i \mathbf{v}_j^\mathsf{T}|i=1,2\dots,n,j=1,2\dots,m\}$, covering all combinations of the left and right singular vectors.
    \end{itemize}
    \label{def: bases}
\end{definition}

\begin{definition}
     The \textbf{direction} of a matrix $\mathbf{A}\in \mathbb{R}^{n \times m}$ ($n<m$) is defined based on its global bases, using an expanded set of its singular values padded with zeros, specifically as $(\sigma_1, 0, \dots, 0, \sigma_2, 0, \dots, 0, \dots, \sigma_n, \dots, 0)\in\mathbb{R}^{nm}$, i.e., flattened $\mathbf{\Sigma}$ by rows.
   \label{def: direction}
\end{definition}
\noindent
Note that any global basis can be regarded as a direction of unit, since its direction is a one-hot vector. For simplicity in further discussions, we will not differentiate between global bases (core bases) and global directions (core directions).

\subsection{What are Task-specific Directions?}\label{sec: pre definition}

We start with the fact that for any specific task, there exists an optimal matrix $\mathbf{W}^*\in\mathbb{R}^{n\times m}$ within the matrix space $\mathbb{R}^{n \times m}$ \cite{si2024see}.
For a pretrained matrix $\mathbf{W}$, its optimal alteration for this specific task is defined as $\Delta \mathbf{W}^* = \mathbf{W}^* - \mathbf{W}$. In PEFT, we only possess the information of $\mathbf{W}$ and the directions of $\mathbf{W}$. Since $\Delta \mathbf{W}^*$ and $\mathbf{W}^*$ are established on their respective bases, we initially project both $\Delta \mathbf{W}^*$ and $\mathbf{W}^*$ onto the global bases of $\mathbf{W}$, capturing their directions.

\begin{definition}
    Define $\mathbf{\Pi}_{\cdot}(\cdot)$ as a projection operator that projects a direction in one coordinate system onto another coordinate system. Specifically, $\mathbf{\Pi}_\mathbf{W}(\mathbf{A})=(p_{11}, \dots, p_{nm})\in\mathbb{R}^{nm}$ is the projection of the direction of a matrix $\mathbf{A}\in\mathbb{R}^{n\times m}$ onto the global bases of another matrix $\mathbf{W}\in\mathbf{R}^{n\times m}$. $p_{ij} = \mathbf{u}_i^\mathsf{T}\mathbf{A}\mathbf{v}_j$ where $\mathbf{u}_i$ and $\mathbf{v}_j$ are the left and right singular vectors of $\mathbf{W}$, respectively.
\end{definition}

\begin{figure*}[ht]
    \centering
    \includegraphics[width=\linewidth]{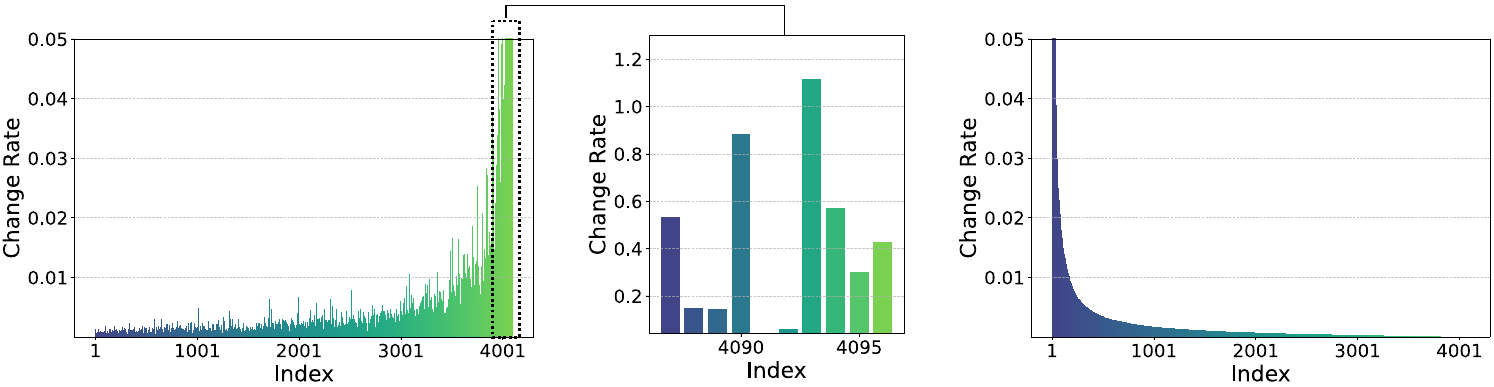}
    \caption{\textbf{Left}: The change rates of $\mathbf{W}$’s core bases, based on $\mathbf{W}^*$, vary significantly across each basis, and the bases with higher change rates tend to be concentrated towards the end. \textbf{Middle}: The change rates for the directions corresponding to the smallest 10 singular values of $\mathbf{W}$. The directions associated with the smallest singular values do not always exhibit the highest change rates. \textbf{Right}: After sorting the change rates from highest to lowest, it is evident that only a few directions have significant change rates, while most exhibit very low change rates. The weights are taken from the 16th layer of LLaMA-7B, and the change rates are scaled (Sec. \ref{sec: supp detail ft LLaMA} for more details). }
    \label{fig:change W star}
\end{figure*}

Established on the global bases of $\mathbf{W}$, $\mathbf{\Pi}_\mathbf{W}(\mathbf{W}^*)$ represents the direction that $\mathbf{W}$ should evolve. $\mathbf{W}$ itself can only utilize up to $n$ bases (i.e., its core bases), meaning that it can at most alter $n$ values of its direction. Thus, we only focus on the changes related to its core directions (for more details, please refer to Sec. \ref{sec: supp why tsd not global}). 
During the transformation, not all core directions' coordinate values change to the same extent. Given the diversity of downstream tasks and the variability in $\mathbf{W}^*$ for different tasks, certain core directions of $\mathbf{W}$ undergo significant changes while others may experience only minimal alterations. Specifically, the change rate $\delta_i$ of the coordinate value corresponding to the $i$-th core direction $\mathbf{u}_i\mathbf{v}_i^\mathsf{T}$ of $\mathbf{W}$ is represented as:
\begin{equation}
\begin{aligned}
    \delta_i &= |\frac{\mathbf{\Pi}_\mathbf{W}(\mathbf{W}^*)_{ii} - \mathbf{\Pi}_\mathbf{W}(\mathbf{W})_{ii}}{\mathbf{\Pi}_\mathbf{W}(\mathbf{W})_{ii}+\epsilon}| = |\frac{\mathbf{u}_i^\mathsf{T}\mathbf{W}^*\mathbf{v}_i - \mathbf{u}_i^\mathsf{T}\mathbf{W}\mathbf{v}_i}{\mathbf{u}_i^\mathsf{T}\mathbf{W}\mathbf{v}_i+\epsilon}| \\
    &=|\frac{\mathbf{u}_i^\mathsf{T}\Delta\mathbf{W}^*\mathbf{v}_i}{\mathbf{u}_i^\mathsf{T}\mathbf{W}\mathbf{v}_i+\epsilon}| = | \frac{\mathbf{\Pi}_\mathbf{W}(\Delta\mathbf{W}^*)_{ii}}{\mathbf{u}_i^\mathsf{T}\mathbf{W}\mathbf{v}_i+\epsilon}|.
\end{aligned}
    \label{eq: delta change rate}
\end{equation}
Here, $\epsilon=10^{-6}$ is a constant to prevent singular values from being zero. $\sigma_i$ is $\mathbf{W}$'s $i$-th singular value, and $\delta_i$ represents the change rate of the coordinate value required in the $i$-th core direction $\mathbf{u}_i \mathbf{v}_i^\mathsf{T}$ for transformation from $\mathbf{W}$ to $\mathbf{\Pi}_\mathbf{W}(\mathbf{W}^*)$, specifically quantifies the extent of adaptation needed for a particular task. We hence define the TSD as:
\begin{definition}
    For a specific task and a pre-trained weight $\mathbf{W}$, considering the optimal weights for this task as $\mathbf{W}^*$, the \textbf{task-specific directions (TSDs)} of this task on $\mathbf{W}$ are $\mathbf{W}$'s core directions whose coordinate values exhibit significantly higher change rates $\delta$ through the alteration from $\mathbf{W}$ to $\mathbf{W}^*$.
    \label{def: tsd}
\end{definition}

Inspired by Eq. (\ref{eq: delta change rate}) and Definition \ref{def: tsd}, it is obvious that $\delta$ is directly related to the projection of $\Delta\mathbf{W}^*$. We are now also equipped to revisit and more precisely articulate LoRA’s conclusions on TSDs:
\begin{itemize}
    \item TSDs are the subset of core directions of pretrained weights $\mathbf{W}$. They are specific to each task, meaning that they vary from one task to another but remain fixed for a given task.
    \item Core directions associated with larger singular values are less likely to be identified as TSDs, as the change rates of their coordinate values are typically smaller than those associated with smaller singular values. It is reasonable since that larger singular values usually encapsulate more generalized information that the model acquired during pretraining. 
\end{itemize}

There are two questions that may arise regarding the definition of TSD as follows:
\subsubsection{Why should TSDs be Established on \texorpdfstring{$\mathbf{W}$}{W}?}
\label{sec: supp tsd W}

The most accurate direction for TSD should be the direction of $\Delta\mathbf{W}^*$. 
However, since we do not have access to $\Delta\mathbf{W}^*$ or $\mathbf{W}^*$ during fine-tuning, we only have information about $\mathbf{W}$. 
Therefore, we establish TSD within the coordinate system based on the directions of $\mathbf{W}$.

\subsubsection{Why Must TSDs be in the Core Bases Rather than Other Global Bases?}\label{sec: supp why tsd not global}
Indeed, other global bases might also experience significant coordinate changes. However, we do not consider them for the following reasons.
First, TSDs are defined based on the change rates. 
Since the coordinate values of $\mathbf{W}$ on the other global bases are zero, even a minimal projection of $\Delta \mathbf{W}$ onto these global bases could result in an excessively high change rate, which would be unreasonable. 
Second, if TSDs were defined based on the absolute magnitude of coordinate changes, consider a scenario where a coordinate value in the core basis and a coordinate value in the global basis change by the same amount. 
If the original coordinate value in the core basis was significantly larger than the change, the change, even if substantial, could be viewed as a minor perturbation relative to the original value. 
Therefore, the same magnitude of change does not necessarily imply that the corresponding directions are equally important. 
For these reasons, we define TSDs based solely on the rate of change and restrict their definition to the core bases.

\subsection{What are the Properties of TSDs?}\label{sec: pre property}

To further explore the properties of TSDs, we fully fine-tune LLaMA-7B \cite{touvron2023llama} on commonsense reasoning tasks, and assume that $\mathbf{W}^*$ can be obtained through fully fine-tuning. 
After acquiring the fully fine-tuned weights $\mathbf{W}^*$, we compute $\Delta\mathbf{W}^* = \mathbf{W}^* - \mathbf{W}$, and based on $\Delta\mathbf{W}^*$, we calculate the change rates of core directions of $\mathbf{W}$. The results are shown in Fig. \ref{fig:change W star}, where we can draw several conclusions (Please refer to Sec. \ref{sec: supp detail ft LLaMA} for more details):
\begin{itemize}
    \item TSD predominantly corresponds to core directions associated with smaller singular values of $\mathbf{W}$, though not the smallest.
    \item TSD encompasses only few directions where substantial change rates occur; most other core directions exhibit minimal or negligible change rates.
\end{itemize}

\begin{figure*}[!ht]
    \centering
    \includegraphics[width=\linewidth]{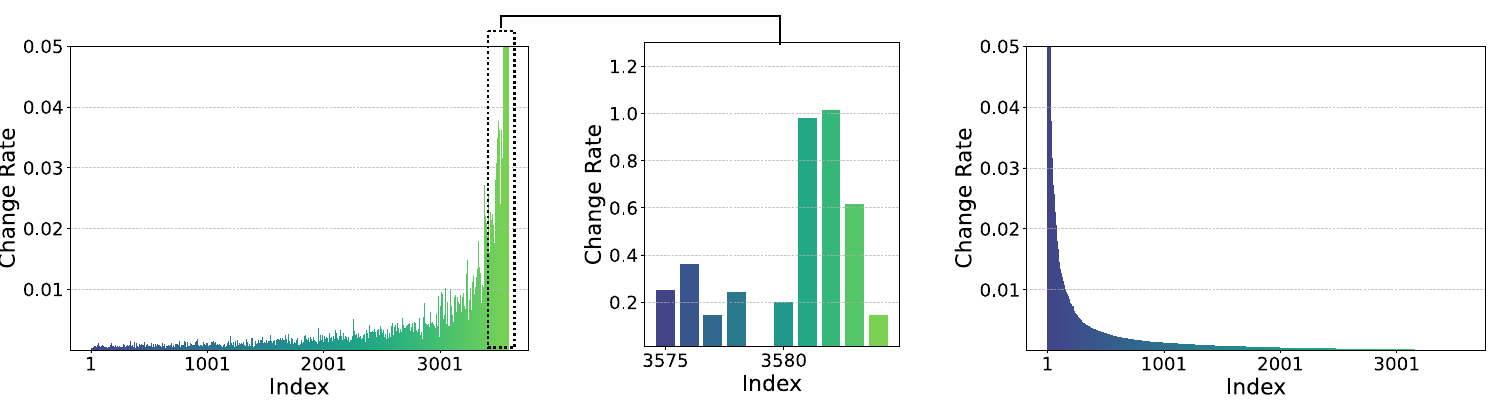}
    \caption{The results when fine-tuning Qwen2.5-7B on math reasoning task are similar to those in Fig. \ref{fig:change W star}}
    \label{fig:change W star qwen}
\end{figure*}

Additionally, it is worth noting that the phenomenon presented in Fig. \ref{fig:change W star} is not an isolated case.
Similar behavior has been observed in other tasks and models. 
For instance, as shown in Fig. \ref{fig:change W star qwen}, results obtained by training Qwen2.5-7B \cite{qwen2.5} on the math reasoning task \cite{hu2023llm} exhibit patterns that are fundamentally consistent with our observations in Fig. \ref{fig:change W star}.

\begin{figure*}[!ht]
    \centering
    \includegraphics[width=\linewidth]{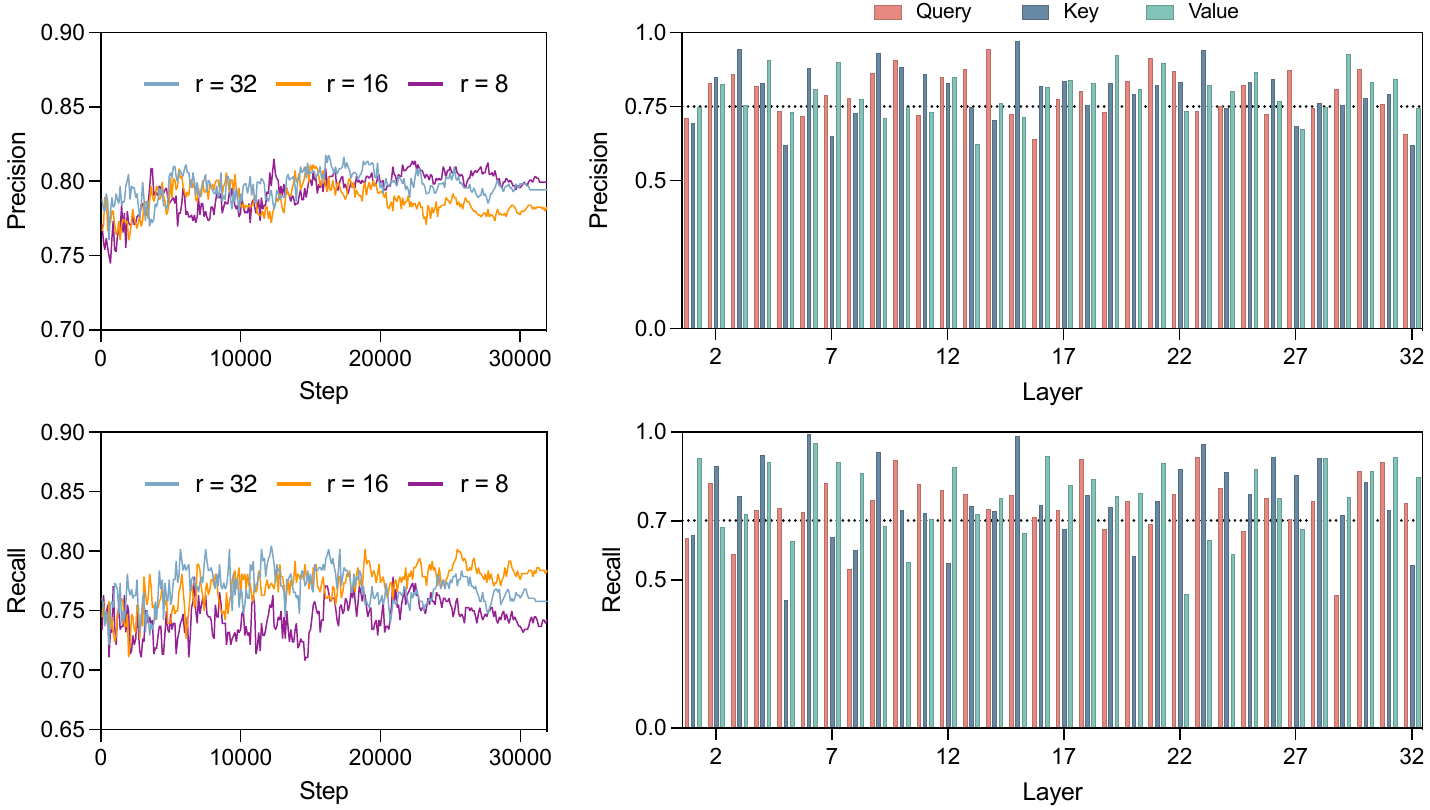}
    \caption{We track the precision and recall of predicted directions every 100 training steps in the query, key and value layers of the LLaMA-7B during LoRA fine-tuning, analyzing how well the continuous updated $\Delta\mathbf{W}$ captures TSDs. \textbf{Left}: We compute the average precision/recall across all query, key, and value layers, showing the model’s ability to retain task-specific knowledge for each training step. Across various rank settings of LoRA, these precision/recall rates consistently exceed 0.75/0.70, indicating that $\Delta\mathbf{W}$ reliably captures and integrates TSD information. \textbf{Right}: For a rank setting of $r=32$, we compute the average precision/recall across all steps for each query, key and value layers, revealing their sensitivity to TSDs. The majority of layers maintain an average precision/recall above 0.75/0.70, showing the robustness to capture TSD information.}
    \label{fig:accuracy TSD lora}
\end{figure*}

\subsection{Challenges and Explorations: Using TSDs is Not as Easy as Expected}\label{sec: met challege}

In downstream tasks, TSDs represent the directions of greatest change, making them critical targets for fine-tuning to adapt pre-trained models effectively to new tasks.
Although we have thoroughly explored the definition and properties of TSDs, a significant challenge is that both $\Delta\mathbf{W}^*$ and $\mathbf{W}^*$ are unknown before fine-tuning, indicating that utilizing TSD information beforehand in practical fine-tuning scenarios is nearly impossible.

Despite the apparent challenges, we place our confidence in the potential of $\Delta\mathbf{W}$. 
We hypothesized that the core directions with the highest change rates predicted by $\Delta\mathbf{W}$ from LoRA are strongly associated with TSDs.
To test this hypothesis, we conducted extensive experiments\footnote{For implementation details, please refer to Sec. \ref{sec: supp detail cr task}.}.
We fine-tune LLaMA-7B using LoRA under three ranks, and for every 100 training steps we record the top 8 core directions of $\mathbf{W}$ which exhibit the highest change rates based on the projection of continuously updated $\Delta\mathbf{W}$ (i.e., the directions related to top 8 largest $\delta_i = \mathbf{u}_i^\mathsf{T} \Delta\mathbf{W} \mathbf{v}_i / \sigma_i$, where $\mathbf{u}_i$, $\mathbf{v}_i$ and $\sigma_i$ are the corresponding left/right singular vectors and values of $\mathbf{W}$). These directions serve as the ``TSDs'' predicted by LoRA. 
We denote these directions as ``launched TSDs'' (LTSDs) for differentiation. We then identify the top 4 or 16 TSDs with the highest change rates by fully finetuning as mentioned in Sec. \ref{sec: pre property}.
We conduct two experimental setups to explore that:
\begin{enumerate}
\item How many of the top 4 TSDs are contained in the 8 LTSDs;
\item How many LTSDs are contained in the top 16 TSDs.
\end{enumerate}
The purpose of the first experiment is more intuitive, aiming to confirm whether the TSDs with the \textit{top highest change rates} can indeed be captured by $\Delta\mathbf{W}$ from LoRA. The second experiment, which may seem a bit more obscure, is designed to address concerns that, besides capturing the top-ranked TSDs, the LTSDs identified might include irrelevant directions. Therefore, this second setup allows us to examine whether the LTSDs predominantly fall within the upper echelons of TSDs based on their change rates, providing insight into the overall relevance and precision of the LTSDs identified by LoRA. Specifically, we employ two metrics to quantify the quality of LTSDs:

\begin{itemize}
\item \textbf{Precision}: Measures how many of the 8 LTSDs are correctly included in the top 16 TSDs. 
\item \textbf{Recall}: Assesses how many of the top 4 TSDs are captured among the 8 LTSDs.
\end{itemize}

The results are presented in Fig. \ref{fig:accuracy TSD lora}, and we also present the statistic results at $r=32$ in Tables. \ref{tab:results across layer}-\ref{tab:results across step}.
It is obvious that the precision and recall of all layers remain consistently high for LoRA across different layers, training steps, and tasks.
Since precision is computed over eight directional classes and recall over four, a mean precision of 75\% corresponds to correctly identifying approximately six out of eight directions, while a recall of 75\% indicates identifying three out of four. 
We also note that the reported standard deviations are relatively large, but this behavior is expected and statistically justified. 
For precision, misclassifying a single direction causes a drop of 12.5\% in the metric, whereas for recall, each misclassification results in a 25\% decrease. 
As the observed standard deviation for precision is close to this 12.5\% unit drop—and the standard deviation for recall remains below the 25\% threshold—we conclude that these variations fall within the normal statistical fluctuation induced by the discrete nature of directional classification.
Moreover, we have also repeated the above experiment by changing the task to the natural language understanding task on the CoLA dataset. 
The results are shown in Tables. \ref{tab:results_across_layer_perturbed}-\ref{tab:results across step 2}. 
These results lead to two significant observations regarding the effectiveness of using $\Delta\mathbf{W}$ to identify TSDs:

\begin{itemize}
\item \textbf{Precision in Identifying Crucial TSDs}: The LTSDs captured by $\Delta\mathbf{W}$ include several of the actual TSDs with the highest change rates, demonstrating that the predicted directions are not only accurate but pinpoint some of the most critical TSDs.
\item \textbf{Overall Coverage of Significant TSDs}: The entirety of the LTSDs resides within the upper echelon of TSDs in terms of change rates. This suggests that while the primary directions identified are indeed the most vital, the remaining directions predicted also belong to the upper tier of important TSDs.
\end{itemize}
	
These insights affirm the utility of $\Delta\mathbf{W}$ in accurately capturing and representing the information of real TSDs. It shows potential in leveraging learned directional changes to identify and utilize key features specific to given tasks, underscoring the robustness of $\Delta\mathbf{W}$ as a predictive tool in practical applications.
It leads to an important conclusion:

\begin{observation}
    Irrespective of the rank setting in LoRA, the training step, or the specific layer within the model, LoRA’s $\Delta\mathbf{W}$ consistently captures the information of the task-specific directions. 
    \label{proposition: lora capture tsd}
\end{observation}

This suggests that even without prior knowledge of TSDs, we can still capture their crucial information through the $\Delta\mathbf{W}$ obtained during LoRA training! 

\section{Unleash the Power of Task-specific Directions} \label{sec: lora-dash}

\begin{figure*}[!ht]
    \centering
    \includegraphics[width=\linewidth]{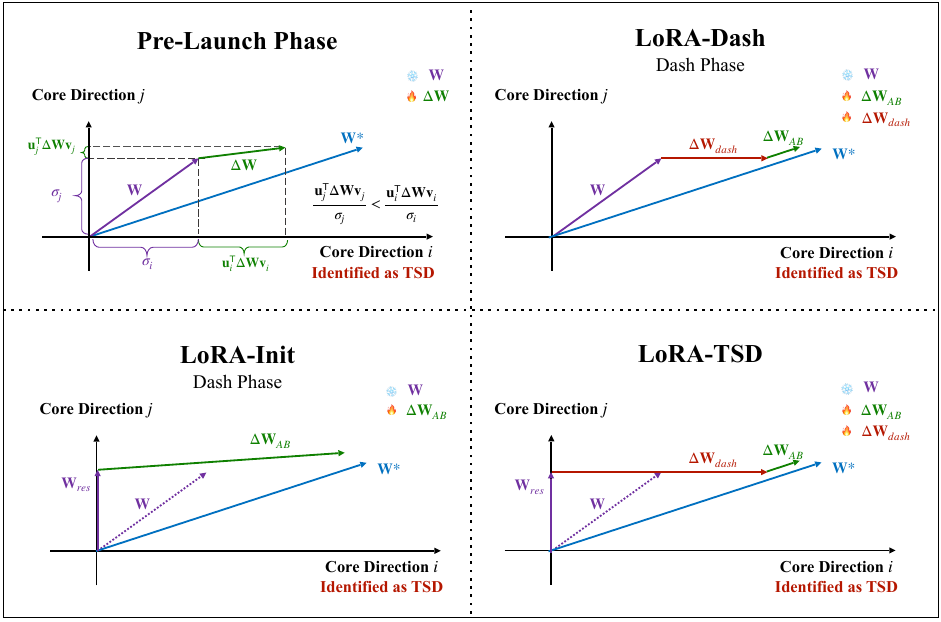}
    \caption{Frameworks of LoRA-TSD, LoRA-Dash and LoRA-Init.}
    \label{fig:dash init framework}
\end{figure*}

To further unleash the power of TSDs in downstream tasks, this section introduces two methods utilizing TSDs.
Both methods are divided into two phases: the ``pre-launch phase'', where TSDs are identified, and the ``dash phase'', which unleash the potential of theses identified TSDs.
Changes occur during the dash phase. Specifically, the first method is LoRA-Dash, which directly simulates the changes in TSDs, while the other, called LoRA-Init, uses TSDs to initialize the matrices $\mathbf{A}$ and $\mathbf{B}$ in LoRA.
The frameworks of LoRA-Dash and LoRA-Init are shown in Fig. \ref{fig:dash init framework}.

\subsection{Pre-Launch Phase} \label{sec: pre-launch phase}

In this phase, LoRA-Dash initially trains the matrices $\mathbf{A}$ and $\mathbf{B}$ (i.e., LoRA’s $\Delta\mathbf{W}$) to capture the information of TSDs. To expedite the utilization of TSD information and based on Proposition \ref{proposition: lora capture tsd}, we posit that there is a predefined number of training steps, $t$, after which $\Delta\mathbf{W}$ can consistently capture TSD information at any stage of training. We set $t$ to 100 as suggested in Sec. \ref{sec: met challege}.

Subsequently, we project $\Delta\mathbf{W}$ onto the core directions of $\mathbf{W}$ to calculate the change rates:
\begin{equation}
    \delta_i = \frac{\mathbf{u}_i^\mathsf{T} \Delta\mathbf{W} \mathbf{v}_i}{\sigma_i + \epsilon},
\end{equation}
where $\mathbf{u}_i$, $\mathbf{v}_i$, and $\sigma_i$ represent the $i$-th left and right singular vectors and singular value of $\mathbf{W}$, respectively.
We identify the top $s$ core directions with the highest change rates, setting $s$ as another hyper-parameter, which is recommended to be 8 as suggested in Sec. \ref{sec: met challege}. We denote $\bar{\mathbf{u}}_i\bar{\mathbf{v}}_i^\mathsf{T}$ for $i = 1, 2, \dots, s$ as the identified ``TSDs'' for dash, To differentiate from the true TSDs derived from $\mathbf{W}^*$ (or theoretically defined), we label these directions as ``launched TSDs'' (LTSDs)\footnote{As shown in Secs. \ref{sec: met challege} and \ref{sec: fur select}, LTSDs and TSDs are heavily overlapped. Given that LTSD closely represents TSD, unless explicitly emphasized otherwise, we treat LTSD as equivalent to TSD for analytical purposes.}.

\subsection{Dash Phase for LoRA-Dash} \label{sec: met dash phase}

To leverage the identified LTSDs, LoRA-Dash directly learns the changes in their coordinates, denoted as $\Delta{\sigma}_i$ for the $i$-th $\bar{\mathbf{u}}_i\bar{\mathbf{v}}_i^\mathsf{T}$, to further fine-tune the model for downstream tasks. $\Delta{\sigma}_i$ is initialized as zero. Mathematically, this adjustment is represented by $\sum_{i=1}^s \Delta{\sigma}_i \bar{\mathbf{u}}_i\bar{\mathbf{v}}_i^\mathsf{T}$.
Ultimately, the updated weight matrix is given by:
\begin{equation}
    \mathbf{W} + \Delta\mathbf{W}_{AB} + \Delta\mathbf{W}_{dash} = \mathbf{W} + \mathbf{A}\mathbf{B} + \sum_{i=1}^s \Delta{\sigma}_i \bar{\mathbf{u}}_i\bar{\mathbf{v}}_i^\mathsf{T}. 
\end{equation}
This equation forms the basis of LoRA-Dash, which aims to fully utilize the power of LTSDs derived from LoRA. During training, $\mathbf{A}$, $\mathbf{B}$, and $\Delta{\sigma}_i$ are continuously updated. 

To further enhance efficiency, we transit to matrix computations for determining change rates.
Let ${\mathbf{U}} =\begin{bmatrix} \mathbf{u}_1 & \ldots & \mathbf{u}_n \end{bmatrix}$ and ${\mathbf{V}} = \begin{bmatrix} \mathbf{v}_1 & \ldots & \mathbf{v}_m \end{bmatrix}$ represent the left and right singular vectors for the weight matrix $\mathbf{W}$, we have
\begin{equation}
\begin{aligned}
     &{\mathbf{U}}^\mathsf{T} \Delta\mathbf{W} \mathbf{V} = \begin{bmatrix}
        \mathbf{u}_1^\mathsf{T} \\ \vdots \\ \mathbf{u}_n^\mathsf{T}
    \end{bmatrix}\Delta\mathbf{W} \begin{bmatrix}
        \mathbf{v}_1 &  \cdots &  \mathbf{v}_m
    \end{bmatrix} \\ &= \begin{bmatrix}
    \mathbf{u}_1^\mathsf{T}\Delta\mathbf{W} \mathbf{v}_1 &  \ldots & \mathbf{u}_1^\mathsf{T}\Delta\mathbf{W} \mathbf{v}_m \\ \vdots & \mathbf{u}_i^\mathsf{T}\Delta\mathbf{W} \mathbf{v}_j & \vdots \\ \mathbf{u}_n^\mathsf{T}\Delta\mathbf{W} \mathbf{v}_1 & \ldots & \mathbf{u}_n^\mathsf{T}\Delta\mathbf{W} \mathbf{v}_m
    \end{bmatrix}.
\end{aligned}
\end{equation}
Therefore, we can derive the change rates $\mathbf{\delta} = \begin{bmatrix}
    \delta_1 & \ldots & \delta_n
\end{bmatrix}^\mathsf{T}$, which are the rectangle diagonal elements of the matrix $(\mathbf{U}^\mathsf{T} \Delta \mathbf{W} \mathbf{V}) \odot (\mathbf{\Sigma}^{-1}+\epsilon\mathbf{I}_{n\times m})$.

\begin{table*}[!ht]
 \renewcommand\arraystretch{1.15}
 \setlength{\tabcolsep}{1.4mm}
    \centering
    \caption{Results on commonsense reasoning tasks. We fine-tune LLaMA-7B \cite{touvron2023llama}, LLaMA2-7B \cite{touvron2023llama2} and LLaMA3-8B \cite{llama3modelcard} on this task.}
    \resizebox{\textwidth}{!}{
    \begin{tabular}{c | l c | c c c  c c c c c | c }
    \toprule
    \textbf{Model} & \textbf{Method} & \textbf{Params(\%)} & \textbf{BoolQ} & \textbf{PIQA} & \textbf{SIQA} & \textbf{HellaS.} & \textbf{WinoG.} & \textbf{ARC-e} & \textbf{ARC-c} & \textbf{OBQA} & \textbf{Avg.} \\
    \toprule
    ChatGPT & - & - & 73.1 & 85.4 & 68.5 & 78.5 & 66.1 & 89.8 & 79.9 & 74.8 & 77.0 \\ \midrule
    
    \multirow{21}{*}{LLaMA-7B} & Fully FT & 100 & 69.9 & 84.2 & 78.9 & 92.3 & 83.3 & 86.6 & 72.8 & 83.4 & 81.4 \\  \cline{2-12}

    & LoRA$_{r=4}$ & 0.10 & 54.7 & 46.1 & 71.5 & 22.4 & 71.6 & 70.9 & 50.1 & 53.4 & 55.1  \\  
    \rowcolor{gray!8}

     \cellcolor{white} & LoRA-Init & 0.10 & 68.4 & 81.3 & 78.1 & 77.9 & 78.8 & 78.8 & 62.8 & 78.4 & 75.6 \\
    \rowcolor{gray!15}
    
     \cellcolor{white} & LoRA-Dash & 0.10 & 65.2 & 79.9 & 78.3 & 82.8 & 77.1 & 78.6 & 65.4 & 78.4 & 75.7 \\ 
     
      \rowcolor{gray!24}
      
   \cellcolor{white} & LoRA-TSD & 0.10 & 66.7 & 80.3 & 78.3 & 81.9 & 78.8 & 79.2 & 64.7 & 79.1 & \textbf{76.1} \\
     
    \cline{2-12}

    & LoRA$_{r=8}$ & 0.21 & 51.4 & 75.3 & 72.0 & 65.1 & 70.1 & 65.8 & 55.7 & 62.6 & 64.8  \\  \rowcolor{gray!8}
    
     \cellcolor{white} & LoRA-Init & 0.21 & 69.5 & 81.9 & 78.1 & 78.8 & 79.8 & 79.2 & 64.9 & 79.2 & 76.2\\
    \rowcolor{gray!15}

    \cellcolor{white} & LoRA-Dash & 0.21 & 69.8 & 81.1 & 77.3 & 85.1 & 81.1 & 77.2 & 64.1 & 79.6 & {76.9} \\

    \rowcolor{gray!24}
      
   \cellcolor{white} & LoRA-TSD & 0.21 & 69.8 & 81.5 & 78.2 & 84.9 & 80.7 & 78.6 & 64.5 & 79.6 & \textbf{77.2} \\
   
    \cline{2-12}
    
    & LoRA$_{r=16}$ & 0.42 & 69.9 & 77.8 & 75.1 & 72.1 & 55.8 & 77.1 & 62.2 & 78.0 & 70.9 \\ 

     & \cellcolor{gray!8}LoRA-Init & \cellcolor{gray!8}0.42 & \cellcolor{gray!8}68.2 & \cellcolor{gray!8}80.3 & \cellcolor{gray!8}77.6 & \cellcolor{gray!8}75.8 & \cellcolor{gray!8}79.3 & \cellcolor{gray!8}77.9 & \cellcolor{gray!8}62.9 & \cellcolor{gray!8}77.4 & \cellcolor{gray!8}74.9 \\

     & \cellcolor{gray!15}LoRA-Dash & \cellcolor{gray!15}0.42 & \cellcolor{gray!15}66.9 & \cellcolor{gray!15}80.2 & \cellcolor{gray!15}77.8 & \cellcolor{gray!15}78.8 & \cellcolor{gray!15}79.2 & \cellcolor{gray!15}78.0 & \cellcolor{gray!15}61.9 & \cellcolor{gray!15}77.4 & \cellcolor{gray!15}75.0 \\ 
     
     \rowcolor{gray!24}
      
   \cellcolor{white} & LoRA-TSD & 0.42 & 69.8 & 80.2 & 78.1 & 78.7 & 79.2 & 78.0 & 62.0 & 78.4 & \textbf{75.6} \\
   
   \cline{2-12}
     
    & LoRA$_{r=32}$ & 0.83 & 68.9 & 80.7 & 77.4 & 78.1 & 78.8 & 77.8 & 61.3 & 74.8 & 74.7 \\  
    
    \rowcolor{gray!8}
    
     \cellcolor{white} & LoRA-Init & 0.83 & 69.4 & 82.3 & 78.1 & 85.4 & 81.3 & 80.9 & 65.7 & 78.6 & 77.7 \\
     
    \rowcolor{gray!15}

    \cellcolor{white} & LoRA-Dash & 0.83 & 69.9 & 82.8 & 78.6 & 84.9 & 81.6 & 82.3 & 66.5 & 80.8 & \textbf{78.4} \\

    \rowcolor{gray!24}
      
   \cellcolor{white} & LoRA-TSD & 0.83 & 69.8 & 82.6 & 78.7 & 85.1 & 81.6 & 81.4 & 66.9 & 81.0 & \textbf{78.4} \\

    \cline{2-12}

    & LoRA$_{r=64}$ & 1.66 & 66.7 & 79.1 & 75.7 & 17.6 & 78.8 & 73.3 & 59.6 & 75.2 & 65.8\\  
    \rowcolor{gray!8}
    
     \cellcolor{white} & LoRA-Init & 1.66 & 68.7 & 81.0 & 78.8 & 81.9 & 81.5 & 79.9 & 65.6 & 79.0 & \textbf{77.0} \\
     
    \rowcolor{gray!15}

    \cellcolor{white} & LoRA-Dash & 1.66 & 69.6 & 79.5 & 76.0 & 82.8 & 75.8 & 81.5 & 64.7 & 81.0 & 76.4  \\

    \rowcolor{gray!24}
      
   \cellcolor{white} & LoRA-TSD & 1.66 & 69.2 & 80.6 & 77.9 & 82.5 & 79.4 & 81.2 & 65.1 & 80.4 & \textbf{77.0} \\

    \midrule

    \multirow{9}{*}{LLaMA2-7B} & Fully FT & 100 & 72.2 & 84.9 & 80.9 & 93.1 & 84.7 & 87.5 & 74.2 & 85.1 & 82.8 \\ \cline{2-12}
    
    & LoRA$_{r=16}$ & 0.41 & 72.4 & 82.5 & 79.1 & 88.5 & 80.6 & 82.3 & 67.1 & 78.9 & 79.3\\ 

    \rowcolor{gray!8}

     \cellcolor{white} & LoRA-Init & 0.41 & 71.7 & 81.6 & 79.5 & 89.5 & 81.9 & 82.9 & 67.9 & 79.6 & 79.3\\

    & \cellcolor{gray!15}LoRA-Dash & \cellcolor{gray!15}0.41 & \cellcolor{gray!15}70.9 &  \cellcolor{gray!15}82.2 & \cellcolor{gray!15}80.5 & \cellcolor{gray!15}90.2 & \cellcolor{gray!15}80.1 & \cellcolor{gray!15}83.5 & \cellcolor{gray!15}68.9 & \cellcolor{gray!15}80.8 & \cellcolor{gray!15}\textbf{79.6}\\ 
    
   & \cellcolor{gray!15}LoRA-TSD & \cellcolor{gray!15}0.41 & \cellcolor{gray!15}71.3 &  \cellcolor{gray!15}81.2 & \cellcolor{gray!15}80.1 & \cellcolor{gray!15}89.9 & \cellcolor{gray!15}81.7 & \cellcolor{gray!15}83.3 & \cellcolor{gray!15}68.4 & \cellcolor{gray!15}80.9 & \cellcolor{gray!15}\textbf{79.6} \\

    \cline{2-12}

    & LoRA$_{r=32}$ & 0.82 & 69.8 & 79.9 & 79.5 & 83.6 & 82.6 & 79.8 & 64.7 & 81.0 & 77.6 \\

    \rowcolor{gray!8}

     \cellcolor{white} & LoRA-Init & 0.82 & 69.8 & 83.0 & 78.0 & 87.7 & 84.1 & 83.1 & 68.3 & 82.0 & {79.5} \\
    
    & \cellcolor{gray!15}LoRA-Dash & \cellcolor{gray!15}0.82 & \cellcolor{gray!15}71.0 & \cellcolor{gray!15}75.7 & \cellcolor{gray!15}79.3 & \cellcolor{gray!15}91.1 & \cellcolor{gray!15}78.6 & \cellcolor{gray!15}84.2 & \cellcolor{gray!15}69.8 & \cellcolor{gray!15}78.8 & \cellcolor{gray!15}78.6 \\

    \rowcolor{gray!24}
      
   \cellcolor{white} & LoRA-TSD & 0.82 & 70.6 & 81.9 & 79.4 & 91.1 & 83.9 & 84.0 & 69.8 & 82.3 & \textbf{80.4} \\

    \midrule

    \multirow{9}{*}{LLaMA3-8B} & Fully FT & 100 & 75.3 & 89.9 & 81.5 & 95.8 & 87.6 & 91.6 & 79.3 & 87.4 & 86.1 \\ \cline{2-12}

    & LoRA$_{r=16}$ & 0.35 & 72.3 & 86.7 & 79.3 & 93.5 & 84.8 & 87.7 & 75.7 & 82.8 & 82.8 \\

    \rowcolor{gray!8}

     \cellcolor{white} & LoRA-Init & 0.35 & 72.2 & 88.3 & 80.9 & 93.3 & 87.0 & 87.7 & 75.8 & 82.8 & 83.4 \\
    
    & \cellcolor{gray!15}LoRA-Dash & \cellcolor{gray!15}0.35 & \cellcolor{gray!15}74.8 & \cellcolor{gray!15}88.0 & \cellcolor{gray!15}80.6 & \cellcolor{gray!15}95.2 & \cellcolor{gray!15}85.6 & \cellcolor{gray!15}89.0 & \cellcolor{gray!15}77.4 & \cellcolor{gray!15}84.8 & \cellcolor{gray!15}84.4 \\

    & \cellcolor{gray!15}LoRA-TSD & \cellcolor{gray!15}0.35 & \cellcolor{gray!15}74.8 &  \cellcolor{gray!15}88.3 & \cellcolor{gray!15}80.9 & \cellcolor{gray!15}95.5 & \cellcolor{gray!15}86.9 & \cellcolor{gray!15}90.4 & \cellcolor{gray!15}78.3 & \cellcolor{gray!15}85.0 & \cellcolor{gray!15}\textbf{85.0} \\
    
    \cline{2-12}

    & LoRA$_{r=32}$ & 0.70 & 70.8 & 85.2 & 79.9 & 91.7 & 84.3 & 84.2 & 71.2 & 79.0 & 80.8 \\ \rowcolor{gray!8}

     \cellcolor{white} & LoRA-Init & 0.70 & 73.6 & 85.5 & 80.0 & 94.6 & 85.7 & 88.6 & 76.1 & 83.0 & 83.4 \\
    \rowcolor{gray!15}

    \cellcolor{white} & LoRA-Dash & 0.70 & 75.3 & 88.5 & 80.2 & 95.7 & 86.8 & 90.7 & 80.2 & 85.6 & 85.4 \\

    \rowcolor{gray!24}
      
   \cellcolor{white} & LoRA-TSD & 0.70 & 75.0 & 88.6 & 80.9 & 96.1 & 87.0 & 89.9 & 80.3 & 85.9 & \textbf{85.5} \\

    \bottomrule
    \end{tabular}}
    \label{tab:results of commonsense}
\end{table*}

\subsection{Dash Phase for LoRA-Init}

In the Dash Phase, in addition to directly simulating the changes in TSDs, we can also use TSDs to initialize the matrices $\mathbf{A}$ and $\mathbf{B}$ in LoRA, which is a practical problem in LoRA series \cite{meng2024pissa, wang2024milora, wang2024loraga}.
Specifically, we perform an SVD decomposition on the pre-trained weights $\mathbf{W}$, which can be split into the top $r$ TSD components and non-TSD components, i.e.,
\begin{equation}
    \mathbf{W} = \sum_{i=1}^r\bar{\sigma}_i\bar{\mathbf{u}}_i\bar{\mathbf{v}}_i^\mathsf{T} + \mathbf{W}_{res}=\bar{\mathbf{U}}\bar{\mathbf{\Sigma}}\bar{\mathbf{V}}^\mathsf{T} + \mathbf{W}_{res}.
\end{equation}
Here, $\bar{\sigma}_i$ represents the corresponding singular value of $\bar{\mathbf{u}}_i\bar{\mathbf{v}}_i^\mathsf{T}$, and $ \mathbf{W}_{res}$ denotes the remaining components, which are frozen during training.
We then use the most significant components that require the greatest changes to initialize the matrices $\mathbf{A}$ and $\mathbf{B}$ as
\begin{equation}
    \mathbf{A}=\bar{\mathbf{U}}\bar{\mathbf{\Sigma}}^{1/2}, \mathbf{B} = \bar{\mathbf{\Sigma}}^{1/2}\bar{\mathbf{V}}^\mathsf{T}.
\end{equation}
Ultimately, the updated weight matrix is given by:
\begin{equation}
    \mathbf{W} \rightarrow \mathbf{W}_{res} + \mathbf{A} \mathbf{B}.
\end{equation}
This equation forms the basis of LoRA-Init. During training, only $\mathbf{A}$ and $\mathbf{B}$ are continuously updated.

\subsection{LoRA-TSD: Unleashing the Power of TSD}

We ultimately combine the two methods to create a fully optimized approach that fully harnesses the power of TSD: LoRA-TSD. 
Specifically, after identifying TSDs, LoRA-TSD both initializes the LoRA matrices $\mathbf{AB}$ using TSDs and directly models the changes in the TSD coordinates.
The update formula for LoRA-TSD is as follows:
\begin{equation}
    \mathbf{W}_{tsd} =\mathbf{W}_{res} +  \mathbf{A} \mathbf{B} +\sum_{i=1}^s \Delta{\sigma}_i \bar{\mathbf{u}}_i\bar{\mathbf{v}}_i^\mathsf{T},
\end{equation}
where $\mathbf{A}$, $\mathbf{B}$ and $\Delta{\sigma}_i $ are trained during fine-tuning. 
LoRA-TSD combines the advantages of both LoRA-Dash and LoRA-Init, fully unleashing the potential of TSD.

\section{How Much Can TSD Boost the Performance of LoRA?} \label{sec: performance}

\begin{table*}[ht]
    \centering
    \renewcommand\arraystretch{1.1} 
    \setlength{\tabcolsep}{3mm}
    \caption {Results with DeBERTaV3 fine-tuned on GLUE development set. ``FT'' represents fully fine-tuning, and ``Base' and ``Large'' represent DeBERTaV3-base and DeBERTaV3-large, respectively.}
    \resizebox{\textwidth}{!}{
    \begin{tabular}{l | c| c c c c c c c c |>{\columncolor{gray!10}}c}
    \toprule
         \multirow{2}{*}{\textbf{Method}} &  \multirow{2}{*}{\textbf{Params(\%)}} & \textbf{MNLI} & \textbf{SST-2} &\textbf{CoLA} & \textbf{QQP} & \textbf{QNLI} & \textbf{RTE} & \textbf{MRPC} & \textbf{STS-B} & \textbf{All}\\
         & & Acc & Acc & Mcc & Acc & Acc & Acc & Acc & Corr & Avg. \\ 
         \midrule
        
        Base(FT) & 100\% & 89.90 & 95.63 & 69.19 & 91.87 & 94.03 & 83.75 & 90.20 & 91.60 & 88.27 \\ \midrule
        
         LoRA$_{r=2}$ & 0.18\% & 90.03 & 93.92 & 69.15 & 90.61 & 93.37 &  87.01 & 90.19 & 90.75 & 88.13  \\  

        \rowcolor{gray!8}
        
        LoRA-Init & 0.18\% & 89.34 & 95.76 & 71.45 & 91.58 & 94.32 &  89.17 & 91.42 & 91.85 & 89.36 \\

        \rowcolor{gray!15}
        
         LoRA-Dash & 0.18\% & 90.14 & 95.42 & 72.41 & 91.65 & 94.36 & 89.89 & 91.67 & 91.64 & {89.65} \\  

         \rowcolor{gray!24}

         LoRA-TSD & 0.18\% & 90.26 & 96.10 & 72.41 & 91.65 & 94.77 & 90.03 & 91.91 & 91.89 & \textbf{89.88} \\

         \midrule

        LoRA$_{r=8}$ & 0.72\% & 89.80 & 93.69 & 69.30 & 91.78 & 92.97 & 86.28 & 90.68 & 91.62 & 88.27  \\  

        \rowcolor{gray!8}
        
        LoRA-Init & 0.72\% & 90.31 & 95.64 & 71.05 & 91.39 & 93.78 & 87.19 & 90.93 & 91.53 & 88.98 \\

        \rowcolor{gray!15}

         LoRA-Dash & 0.72\% & 90.55 & 96.99 & 70.78 & 92.39 & 94.22 & 88.09 & 91.18 & 91.91 & 89.52  \\  

        \rowcolor{gray!24}

         LoRA-TSD & 0.72\% & 90.37 & 96.70 & 72.41 & 93.44 & 94.77 & 90.10 & 91.91 & 92.07 & \textbf{90.22} \\

         \bottomrule \toprule

         Large(FT) & 100\% & 91.81 & 96.93 & 75.27 & 93.01 & 96.02 & 92.68 & 92.20 & 92.98 & 91.36 \\ \midrule
        
         LoRA$_{r=2}$ & 0.20\% & 91.33 & 95.87 & 73.89 & 91.84 & 95.14 & 91.69 & 90.68 & 92.85 & 90.41 \\  

         \rowcolor{gray!8}
        
        LoRA-Init & 0.20\% & 91.59 & 96.10 & 75.34 & 92.67 & 95.48 & 92.06 & 91.67 & 92.96 & 90.98 \\

        \rowcolor{gray!15}

         LoRA-Dash & 0.20\% & 91.65 & 96.11 & 76.11 & 92.61 & 95.52 & 92.78 & 92.18 & 93.05 & 91.25  \\  

         \rowcolor{gray!24}

         LoRA-TSD & 0.20\% & 91.65 & 96.28 & 76.34 & 92.89 & 95.77 & 92.93 & 92.03 & 92.89 & \textbf{91.35} \\
         
         \midrule

        LoRA$_{r=8}$ & 0.80\% & 91.38 & 96.33 & 74.48 & 92.54 & 95.48 & 92.05 & 91.17 & 92.92 & 90.79 \\  

        \rowcolor{gray!8}
        
        LoRA-Init & 0.80\% & 91.30 & 96.44 & 76.34 & 92.49 & 95.75 & 92.42 & 91.67 & 92.73 & 91.14 \\

        \rowcolor{gray!15}

         LoRA-Dash & 0.80\% & 91.18 & 96.45 & 76.88 & 92.81 & 95.85 & 93.51 & 91.91 & 92.86 & 91.43 \\  

         \rowcolor{gray!24}

         LoRA-TSD & 0.80\% & 91.27 & 96.51 & 76.94 & 92.99 & 95.78 & 93.51 & 92.18 & 93.05 & \textbf{91.53} \\

        \bottomrule
    \end{tabular}
    }
    \label{tab: deberta results}
\end{table*}

To explore the performance gains of LoRA-Dash, LoRA-Init, and LoRA-TSD, both numerical and visual comparisons are employed. 
The numerical experiments focus on commonsense reasoning and natural language understanding tasks, assessing performance improvements of our methods in standard natural language processing metrics. 
Visual comparisons focuses on subject-driven generation task, involving using a text-to-image diffusion model to learn specific concepts (i.e., task-specific concept), providing a clear, qualitative and intuitive understanding of the impact of TSDs on performance. 

\subsection{Implementation Details}\label{sec: supp detail imple}
We compare our methods mainly with LoRA. The ranks of LoRA-Dash, LoRA-Init, LoRA-TSD, LoRA and other methods are varied among \{2, 4, 8, 16, 32, 64\}.
All methods are implemented using the publicly available PyTorch \cite{paszke2019pytorch} framework, and all experiments are conducted on NVIDIA A100 GPUs\footnote{We have also test that most of the experiments can also be conducted on one consumer GPU resource such as NVIDIA RTX3090.}. 
The hyper-parameter $t$ is set to 100, and $s=8$ for LoRA-TSD and LoRA-Dash. 
We fine-tune all the layers of different models in all the experiments, and the detailed experimental settings are shown in Secs. \ref{sec: supp detail cr task}-\ref{sec: supp detail sdg task}, respectively.

\subsection{Numerical Results}

The numerical results, as shown in Tables. \ref{tab:results of commonsense}-\ref{tab: deberta results}, demonstrate that LoRA-Dash, LoRA-Init and LoRA-TSD significantly outperform LoRA across all tasks and models, with LoRA-TSD achieving the best performance.
This distinction is so pronounced that further elaboration on performance gains might be unnecessary. 
Therefore, we instead focus on several key insights derived beyond mere performance gains. 

\textbf{Robustness to Parameter Budget:} (When fine-tuning LLaMA-7B) LoRA shows enhanced performance with an increased parameter budget, indicating its dependence on larger rank sizes.
Conversely, our methods maintain strong performance even under limited parameter conditions, which suggests that TSDs are crucial for maximizing fine-tuning efficiency.
This highlights its effectiveness in harnessing the full potential of TSDs, even in constrained settings.

\textbf{Surpassing Fully Fine-Tuning Performance:}
While LoRA typically falls short of the results achieved by fully fine-tuning, our methods in some cases outperform fully fine-tuning.
This intriguing observation may provide new insights into optimizing training strategies for fine-tuning. 

For comparisons with other methods, please refer to Tables. \ref{tab:results of commonsense compare others}-\ref{tab: deberta results comparison}.

\subsection{Visual Results}

\begin{figure*}[!ht]
    \centering
    \includegraphics[width=\linewidth]{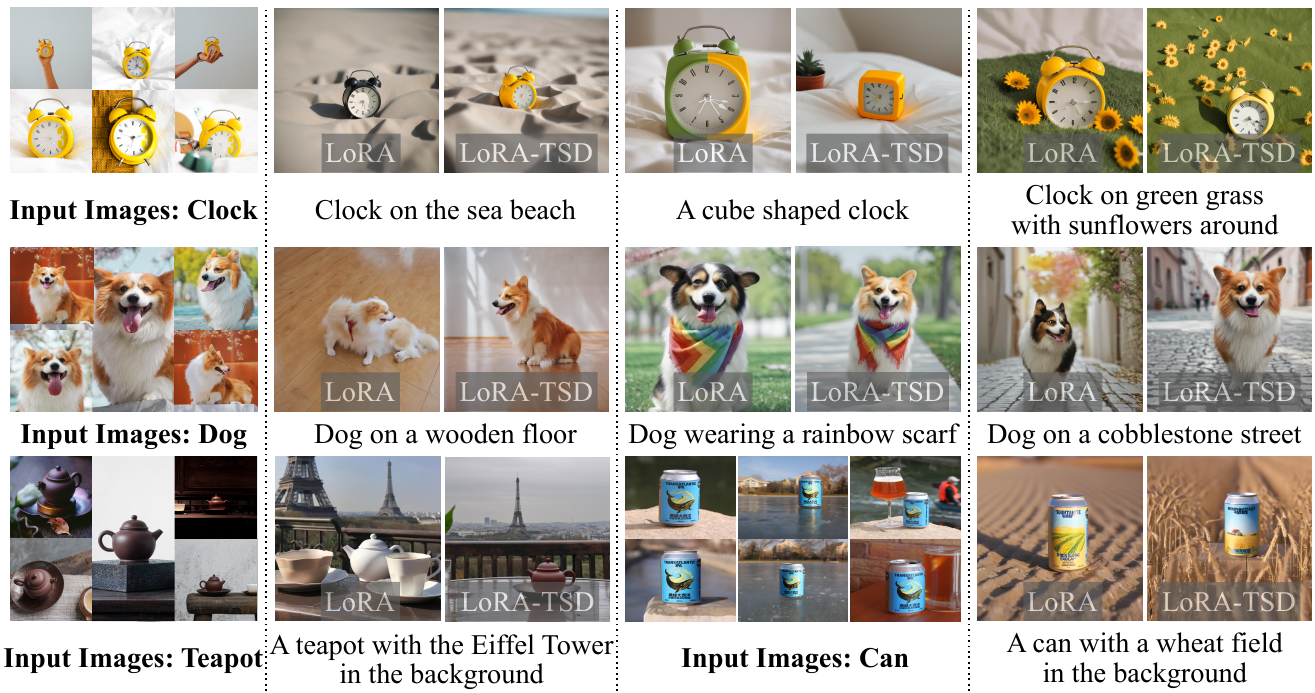}
    \caption{Comparison of generated images from LoRA and LoRA-TSD on subject-driven generation task. Our method consistently aligns more closely with the subjects in the input images and adheres better to the given prompts than LoRA.}
    \label{fig:visual comparison}
\end{figure*}

\begin{figure*}[!ht]
    \centering
    \includegraphics[width=\linewidth]{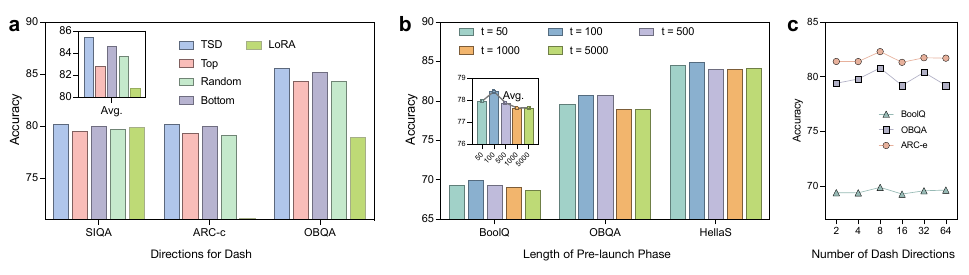}
    \caption{Ablation study of LoRA-Dash. \textbf{a}. Ablation study on the directions for dash. \textbf{b}. Ablation study on the length of pre-launch phase $t$. \textbf{c}. Ablation study on the number of directions $s$ for dash.}
    \label{fig: ablation}
\end{figure*}

\begin{figure*}
    \centering
    \includegraphics[width=\textwidth]{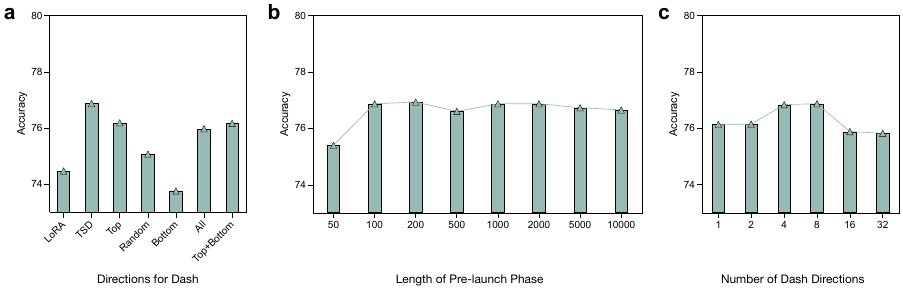}
    \caption{Ablation study on CoLA when fine-tuning DeBERTaV3-large of LoRA-Dash. \textbf{a}. Ablation study on the directions for dash. Beyond the top, random, and bottom directions, we also explored adjusting all directions and a combined adjustment of both bottom and top directions.  \textbf{b}. Ablation study on the length of pre-launch phase $t$. \textbf{c}. Ablation study on the number of directions $s$ for dash.}
    \label{fig: ablation more}
\end{figure*}

\begin{table}[!ht]
\centering
\setlength{\tabcolsep}{4mm}
\caption{Performance comparison between LoRA and LoRA-TSD on CLIP text-image alignment and image similarity metrics.}
\label{tab:clip_scores}
\resizebox{\linewidth}{!}{
\begin{tabular}{lcc}
\toprule
\textbf{Metric (\%)} & LoRA & LoRA-TSD \\
\midrule
Mean CLIP Text-Image Score & 32.0 & 33.1 \\
Mean Image Similarity & 76.4  & 78.6 \\
\bottomrule
\end{tabular}}
\end{table}

The visual comparisons, as shown in Fig. \ref{fig:visual comparison}, demonstrate the superior fidelity of images generated by our method in aligning with the subjects of the input images compared to those by standard LoRA.
For instance, images generated by LoRA of a dog and a teapot showed significant deviations from the input, whereas our method's outputs maintained high consistency with the original images.
Additionally, the effectiveness of TSDs is particularly evident in the response of our method to given prompts. 
TSDs effectively encapsulate the specific information of each prompt, enabling our methods to precisely grasp the semantic essence of prompts such as \textit{rainbow scarf} or \textit{wheat field}.
This capability allows LoRA-TSD to adeptly render complex themes and details within the images.

We also conducted a user study using the DreamBooth dataset. 
We selected 10 different categories and generated corresponding images using five prompts for each category. 
We then invited 36 experts to evaluate the images based on subject similarity and prompt consistency, with each expert evaluating randomly sampled 25 pairs of images.
During evaluation, the pairs of images were shuffled, and the experts were unaware of the source model for each image. 
Overall, the images LoRA-TSD generated achieved an 88.11\% approval rating. 
This suggests that the images generated by LoRA-TSD were more favorably recognized by experts w.r.t. subject similarity and prompt consistency.

Moreover, we also evaluated the performance using the text-image score evaluated by CLIP \cite{radford2021learning} and the image similarity score. 
The CLIP text-image score measures the alignment between textual descriptions and the corresponding images, while the image similarity score quantifies the similarity between the original image and the generated image. 
The results are presented in Table \ref{tab:clip_scores}, where it is evident that LoRA-TSD outperforms LoRA on both metrics. 
The images generated by LoRA-TSD exhibit better text alignment and a stronger correspondence with the original images, demonstrating that LoRA-TSD achieves superior results compared to LoRA.

\subsection{Ablation Study and Sensitivity Analysis of LoRA-Dash}
\label{sec: fur dash num and length}

We first investigated the effects of TSDs during the pre-launch phase compared to other locations, such as selecting top, bottom or random core directions. The results when fine-tuning LLaMA3-8B and DeBERTaV3-large, as shown in Figs. \ref{fig: ablation}(a) and \ref{fig: ablation more}(a), clearly indicate that selecting TSDs yields the best performance, with top selections performing the worst. This is not difficult to understand: directly dashing TSDs maximizes the ability of $\mathbf{W}$ to adapt to downstream tasks.
Since TSDs are generally located towards the end of the spectrum, choosing directions from the bottom naturally results in better outcomes than selecting from the top.

We then evaluate the influence of two hyper-parameters $t$ and $s$ in LoRA-Dash when fine-tuning LLaMA-7B and DeBERTaV3-large, and the results are illustrated in Figs. \ref{fig: ablation}(b)-(c) and \ref{fig: ablation more}(b)-(c).
It can be observed that starting the dash phase after a longer pre-launch period tends to yield slightly inferior results compared to starting it earlier. 
This could be attributed to the fact that entering the dash phase earlier allows for more extensive utilization of TSDs, potentially enhancing the adaptation to downstream tasks. 
Additionally, a larger value of $s$, representing the number of directions used in the dash phase, also impacts performance. 
Based on our following analysis in Sec. \ref{sec: fur amplify}, we suspect that the inclusion of some irrelevant directions might introduce noise into the training process (i.e., destabilizing model convergence), leading to diminished performance.

Based on extensive empirical evaluations across multiple tasks, we suggest the following guidelines.
First, the transition into the Dash phase should occur once training has reached a stable regime; in practice, this happens quickly, and we therefore recommend setting $t$ to a relatively small value, with a default of 100 steps, which is sufficient for convergence stabilization in most settings.
Second, regarding the number of directions $s$, we use a default of 8. 
More complex tasks generally benefit from a larger number of directions, while simpler tasks tend to prefer fewer. 

\subsection{Ablation Study of LoRA-Init}
We also conducted ablation experiments on LoRA-Init.
We initialized the matrices $\mathbf{A}$ and $\mathbf{B}$ with different methods, such as using the top singular vectors of $\mathbf{W}$ (PISSA \cite{meng2024pissa}), the bottom singular vectors (MiLoRA \cite{wang2024milora}), or random singular vectors (RoSA \cite{hameed2024rosa}), and fine-tuned LLaMA3-8B using LoRA. 
The experimental results, as shown in Table \ref{tab:ablation lora-init}, clearly demonstrate that initializing LoRA with TSDs yields better results, providing evidence for the effectiveness of TSDs.

\begin{table*}[!ht]
 \renewcommand\arraystretch{1}
 \setlength{\tabcolsep}{2.5mm}
    \centering
    \caption{Ablation study of LoRA-Init when fine-tuning LLaMA3-8B on commonsense reasoning tasks.}
    \resizebox{\textwidth}{!}{
    \begin{tabular}{ c c | c c c  c c c c c | c }
    \toprule
     \textbf{Method} & \textbf{Params} & \textbf{BoolQ} & \textbf{PIQA} & \textbf{SIQA} & \textbf{HellaS.} & \textbf{WinoG.} & \textbf{ARC-e} & \textbf{ARC-c} & \textbf{OBQA} & \textbf{Avg.} \\
    \toprule

    LoRA$_{r=32}$ & 0.70\% & 70.8 & 85.2 & 79.9 & 91.7 & 84.3 & 84.2 & 71.2 & 79.0 & 80.8 \\

    PISSA & 0.70\% & 67.1 & 81.1 & 77.2 & 83.6 & 78.9 & 77.7 & 63.2 & 74.6 & 75.4\\

    MiLoRA & 0.70\% & 68.8 & 86.7 & 77.2 & 92.9 & 85.6 & 86.8 & 75.5 & 81.8 & 81.9 \\

    RoSA & 0.70\% & 68.9 & 84.2 & 77.2 & 88.4 & 82.5 & 81.7 & 71.1 & 80.2 & 79.3 \\
    
    \rowcolor{gray!15}

    LoRA-Init & 0.70\% & 73.6 & 85.5 & 80.0 & 94.6 & 85.7 & 88.6 & 76.1 & 83.0 & 83.4 \\
    
    \bottomrule
    \end{tabular}}
    \label{tab:ablation lora-init}
\end{table*}

\subsection{Convergence Curve}

\begin{figure*}[!ht]
    \centering
    \includegraphics[width=\linewidth]{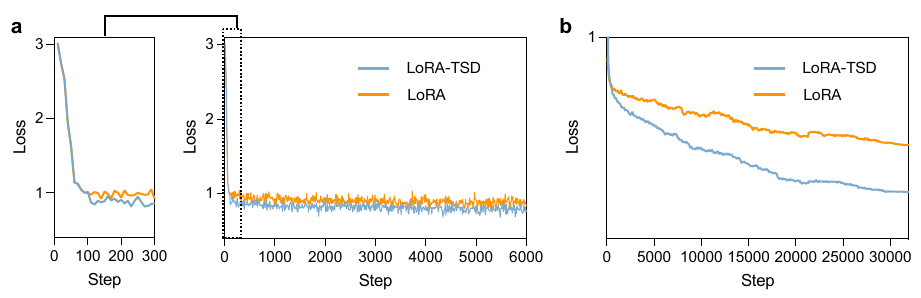}
    \caption{\textbf{a}. Training loss when fine-tuning LLaMA-7B. We record the training loss every 10 steps. \textbf{b}. Validation loss when fine-tuning LLaMA-7B. We record the training loss every 80 steps.}
    \label{fig:loss curve}
\end{figure*}

We present loss curves for LoRA and one of our method LoRA-TSD when fine-tuning LLaMA-7B. 
We ensure consistency in all training hyper-parameters, and the results are shown in Fig. \ref{fig:loss curve}(a). 
We can observe that during the pre-launch phase, the loss trajectories of LoRA-TSD and LoRA are aligned. 
However, as the training progresses into the dash phase, LoRA-TSD consistently exhibits a slightly lower loss compared to standard LoRA. 

One may concern that a lower training loss could lead to overfitting.
Therefore, we also tracked the validation loss throughout the training process, with the results presented in Fig. \ref{fig:loss curve}(b). 
It is clear that the validation loss for LoRA-TSD is much lower than that for LoRA.
This suggests that LoRA-TSD not only avoids overfitting but also enhances its learning efficacy by incorporating task-specific knowledge.

\begin{table*}[ht]
    \renewcommand\arraystretch{1}
    \setlength{\tabcolsep}{2mm}
    \centering
    \caption{Alignment on the directions of LoRA-TSD and TSDs. We report the percentage from the self-attention value (v) projection of the first, middle and last layer of two models, as well as the average percentage of all q, k, and v modules.}
    \resizebox{\textwidth}{!}{
    \begin{tabular}{c | c c c c | c c c c}

    \toprule
        Model & \multicolumn{4}{c|}{LLaMA-7B ($r=8$)} &  \multicolumn{4}{c}{DeBERTaV3-large ($r=2$)} \\
         
        \midrule
        
        Layer & First (1st) & Middle (16th) & Last (32nd) & All. & First (1st) & Middle (12nd) & Last (24th) & All.  \\
         \midrule
         DTSDs $\cap$ LTSDs & 1.00 & 1.00 & 1.00 & 0.76 & 1.00 & 1.00 & 1.00 & 0.99
         
         \\ \midrule
         
         TSDs $\cap$ LTSDs & 1.00 & 1.00 & 1.00 & 0.78 & 0.75 & 0.75 & 0.50 & 0.73
         
         \\ \midrule

         TSDs $\cap$ DTSDs & 1.00 & 1.00 & 1.00 & 0.85 & 0.75 & 0.75 & 0.75 & 0.82
         
         \\

    \midrule
    \midrule
    
         Model & \multicolumn{4}{c|}{LLaMA-7B ($r=16$)} & \multicolumn{4}{c}{DeBERTaV3-large ($r=8$)} \\
         
        \midrule
        
         Layer & First (1st) & Middle (16th) & Last (32nd) & All. & First (1st) & Middle (12nd) & Last (24th) & All.  \\
         \midrule
         DTSDs $\cap$ LTSDs &  1.00 & 1.00 & 0.75 & 0.75 & 1.00 & 1.00 & 1.00 & 0.98
         \\ \midrule
         
         TSDs $\cap$ LTSDs & 0.75 & 0.75 & 1.00 & 0.74 & 1.00 & 1.00 & 1.00
         & 0.84
          
         \\ \midrule

         TSDs $\cap$ DTSDs & 1.00 & 0.75 & 1.00 & 0.86 & 1.00 & 1.00 & 1.00 & 0.91
         
         \\

    \midrule
    \midrule
         Model & \multicolumn{4}{c|}{LLaMA-7B ($r=32$)} &  \multicolumn{4}{c}{DeBERTaV3-large ($r=32$)} \\
         
        \midrule
        
         Layer & First (1st) & Middle (16th) & Last (32nd) & All. & First (1st) & Middle (12nd) & Last (24th) & All.  \\
         \midrule
         DTSDs $\cap$ LTSDs & 
        1.00 & 0.75 & 0.75 & 0.76 & 0.75 &     1.00 & 1.00 & 0.96
         \\ \midrule
         
         TSDs $\cap$ LTSDs & 0.75 & 0.75 & 0.75 & 0.77 & 0.75 & 0.75 & 1.00 & 0.82
         \\ \midrule

         TSDs $\cap$ DTSDs & 1.00 & 1.00 & 1.00 & 0.81 & 0.75 & 1.00 & 1.00 & 0.79
         
         \\
         \bottomrule
    \end{tabular}}
    \label{tab:tsd alignment}
\end{table*}

\section{Analysis of Task-specific Directions} \label{sec: understanding lora-dash}

In this section, we conduct extensive experiments to better understand task-specific directions, with the method LoRA-TSD for illustration.
We first explore the relationships between the LTSDs, the final directions after the dash phase, and TSDs to determine whether LoRA-TSD can amplify the information of TSDs (Sec. \ref{sec: fur select}) and, if so, to what extent (Sec. \ref{sec: fur amplify}).
We then explore whether TSD are ``task-specific'' (Sec. \ref{sec tsd task-specific}), and verify whether the information in TSDs can help enhance the effectiveness of other methods (Sec. \ref{sec: fur tsd help other}). Finally, 
We provide more detailed discussions on the task-specific directions.

\subsection{How Do TSDs Align with the Final Directions?}\label{sec: fur select}

We examine the alignment between the LTSDs identified during the pre-launch phase, the directions with highest change rates pinpointed by the final weight change of LoRA-TSD (denoted as ``DTSDs'', Delta TSDs), and the true TSDs derived from $\mathbf{W}^*$. Following the settings in Secs. \ref{sec: met challege}, we report the percentage that: 
\begin{itemize}
    \item How many of the top 4 DTSDs directions are contained in $s$ LTSDs\footnote{Here since $s=8$, the setting is equal to that in Sec. \ref{sec: met challege}.}. We denote this percentage as ``DTSDs $\cap$ LTSDs '' for further convenience.
    
    \item How many of the top 4 TSDs are contained in $s$ LTSDs. Denote this as ``TSDs $\cap$ LTSDs''
    
    \item  How many of the top 4 TSDs are in $s$ DTSDs. Denote this as ``TSDs $\cap$ DTSDs''.
\end{itemize}
The first is to explore the consistency between the directions selected during the pre-launch phase and those showing the greatest change after the final training, and the second and third aim to validate the consistency between the directions chosen during the pre-launch phase and the final trained with the true TSDs.
We conducted tests on LLaMA-7B \cite{touvron2023llama} and DeBERTaV3-large \cite{he2021debertav3}, each with three ranks of LoRA-TSD, and the results are shown in Table. \ref{tab:tsd alignment}. 
It is obvious that LoRA-TSD indeed captures a great proportion of TSDs' information. Moreover, we can also obtain two significant findings: First, the LTSDs and DTSDs of LoRA-TSD align closely, which indicates that \textit{\textbf{the directions amplified by LoRA-TSD finally are essentially consistent with those selected during the pre-launch phase}}.
Therefore, since LTSDs contain substantial information about TSDs, it is not surprising that the final DTSDs consistently include information about the true TSDs.
Second, when compared with the recall results in Fig. \ref{fig:accuracy TSD lora} where the end of each line indicates the final percentage of TSDs $\cap$ DTSDs of LoRA, it becomes evident that \textit{\textbf{LoRA-TSD’s trained weights capture a greater proportion of TSD information than vanilla LoRA}}.

         
        



        

\subsection{How much can TSDs be Amplified?}\label{sec: fur amplify}
Since LoRA-TSD captures the information of TSDs, we further explore to what extent LoRA-TSD can amplify the information of these directions. 
To quantify this amplification, we define an amplification factor that measures how significantly the task-specific features are enhanced. 
We first project the pretrained weights $\mathbf{W}$ and the weights merged after training, i.e., $\mathbf{W}_{res}+\Delta\mathbf{W}_{all}$, onto the LTSDs. 
Let $\bar{\mathbf{U}} = \begin{bmatrix}\bar{\mathbf{u}}_1 & \ldots & \bar{\mathbf{u}}_s\end{bmatrix}$ and $\bar{\mathbf{V}} = \begin{bmatrix}\bar{\mathbf{v}}_1 & \ldots & \bar{\mathbf{v}}_s\end{bmatrix}$, we can then define the amplification factor of $\Delta\mathbf{W}_{all}$ as the Frobenius norms of these two weights on these directions as
$ \frac{\|\bar{\mathbf{U}}^\mathsf{T} \mathbf{W}_{tsd}\bar{\mathbf{V}}\|_F}{\|\bar{\mathbf{U}}^\mathsf{T} \mathbf{W} \bar{\mathbf{V}}\|_F}$.
Specifically, we also evaluate the separate impacts of $\Delta\mathbf{W}_{AB}$ and $\Delta\mathbf{W}_{dash}$ in the amplification, with the corresponding amplification factor being $\frac{\|\bar{\mathbf{U}}^\mathsf{T} (\mathbf{W}_{res} + \Delta\mathbf{W}_{AB} )\bar{\mathbf{V}}\|_F}{\|\bar{\mathbf{U}}^\mathsf{T} \mathbf{W} \bar{\mathbf{V}}\|_F}$ and $\frac{\|\bar{\mathbf{U}}^\mathsf{T} (\mathbf{W}_{res} + \Delta\mathbf{W}_{dash} )\bar{\mathbf{V}}\|_F}{\|\bar{\mathbf{U}}^\mathsf{T} \mathbf{W} \bar{\mathbf{V}}\|_F}$.

We conduct this experiment on LLaMA-7B with three ranks. 
The results shown in Table. \ref{tab:tsd amplification} demonstrate that \textbf{\textit{LoRA-TSD significantly enhances the features associated with LTSDs under all test conditions}}.
Notably, the feature amplification contributed by $\Delta\mathbf{W}_{dash}$ accounts for the majority of the overall enhancement, whereas the amplification from $\Delta\mathbf{W}_{AB}$ is relatively minor in comparison. 
\textbf{\textit{LoRA-TSD effectively learns and amplifies the majority of TSD information with only $s$ parameters (i.e., $\Delta\sigma_i$ in $\Delta\mathbf{W}_{dash}$), reducing the learning burden on $\Delta\mathbf{W}_{AB}$ during fine-tuning}}. 
This efficiency also allows $\Delta\mathbf{W}_{AB}$ to potentially focus on learning other significant features, further optimizing the model’s performance on downstream tasks.

\begin{table*}[ht]
    \renewcommand\arraystretch{0.8}
    \setlength{\tabcolsep}{2.5mm}
    \centering
    \caption{Amplification on task-specific features of LoRA-TSD. We report the amplification factor from the self-attention key (k) projection of the first, middle, and last layers of LLaMA-7B, as well as the average of all q, k, and v modules. }
    \resizebox{\textwidth}{!}{
    \begin{tabular}{c | c c c | c c c | c c c }
    \toprule
         \multirow{2}{*}{Layer} & \multicolumn{3}{c|}{$r=8$} &  \multicolumn{3}{c|}{$r=16$} & \multicolumn{3}{c}{$r=32$} \\

        & $\Delta\mathbf{W}_{all}$ & $\Delta\mathbf{W}_{AB}$ & $\Delta\mathbf{W}_{dash}$ & $\Delta\mathbf{W}_{all}$ & $\Delta\mathbf{W}_{AB}$ & $\Delta\mathbf{W}_{dash}$ & $\Delta\mathbf{W}_{all}$ & $\Delta\mathbf{W}_{AB}$ & $\Delta\mathbf{W}_{dash}$ \\ \midrule
        
         1st & 285.12 & 45.32 & 280.05 & 635.22 & 173.58 & 595.75 & 45.50 & 15.62 & 40.20 \\ \midrule

         16th & 12.50 & 5.10 & 10.02 & 14.25 & 8.12 & 12.01 & 2.35 & 1.72 & 2.10 \\ \midrule

         32nd & 70.45 & 8.00 & 68.02 & 32.10 & 12.75 & 30.40 & 8.85 & 1.80 & 8.75 \\ \midrule

         Avg. & 23.56 & 7.88 & 22.37 & 33.52 & 13.82 & 30.04 & 5.57 & 2.71 & 5.35 \\
        
         \bottomrule
    \end{tabular}}
    \label{tab:tsd amplification}
\end{table*}

        



        

\begin{figure*}[!ht]
    \centering
    \includegraphics[width=\linewidth]{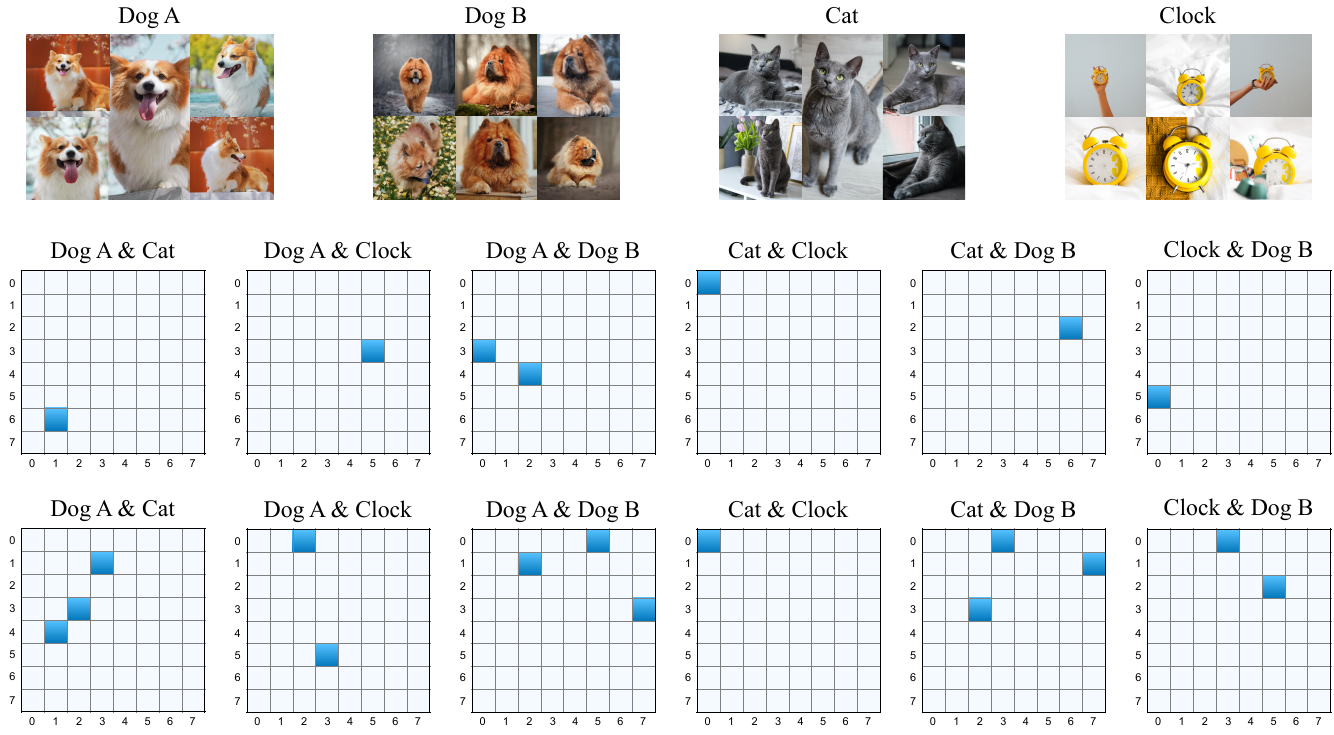}
    \caption{The results of the TSDs similarity between different tasks (subjects). For the first row: Images from different subjects. The second and third rows contain results extracted from two different layers of U-net \cite{ronneberger2015u}. Each subplot illustrates the similarity of TSDs captured for two different subjects. The title of each subplot indicates the respective subjects, with the directional order of the first subject represented by columns and the second by rows. The deep blue blocks highlight where the directions for the two subjects coincide.}
    \label{fig:tsd true ts}
\end{figure*}

\subsection{Are TSDs ``Task-specific''?}\label{sec tsd task-specific}

We sought to verify whether TSDs are genuinely task-specific, that is, whether the most critical directions differ for each task. To investigate this, we employed a prototypical example: the subject-driven generation task. In this setting, we tracked the TSDs captured by LoRA-TSD when different subjects served as targets and analyzed the relationships between these TSDs. The results are presented in Fig. \ref{fig:tsd true ts}.

The experimental results affirm the task-specific nature of TSDs; indeed, each task exhibits distinct TSDs with minimal overlap between tasks. 
Moreover, even when tasks share the same direction, its importance varies significantly between them.
For instance, as depicted in the central subfigure of the first column in Fig. \ref{fig:tsd true ts}, both ``Dog A'' and ``Cat'' share a common direction, yet it ranks as the seventh most significant for ``Dog A'' and the second most significant for ``Cat''. 
Furthermore, tasks that are closely related tend to share more directions, as observed between ``Dog A'' and  ``Dog B''; however, the importance of these shared directions still differs for each task. This variability underscores the fundamentally task-specific property of TSDs, illustrating their unique and variable impact across different contexts.

\subsection{Can TSDs Enhance Other Methods' Performance?}\label{sec: fur tsd help other}

To investigate whether TSDs are beneficial for other PEFT methods, we extended our experiments to include AdaLoRA \cite{zhang2022adaptive} and FLoRA \cite{si2024flora}, assessing the impact of leveraging TSDs within these frameworks. The results, as shown in Table. \ref{tab: flora and adalora of commonsense}, clearly indicate that incorporating TSDs significantly enhances the performance of these methods as well. This underscores the pivotal role of TSDs in optimizing model behavior for downstream tasks, highlighting their universal applicability across various PEFT strategies.

\begin{table*}[!ht]
 \renewcommand\arraystretch{1}
    \centering
    \caption{Results of commonsense reasoning tasks when fine-tuning LLaMA-7B.}
    \resizebox{\textwidth}{!}{
    \begin{tabular}{c c | c c c  c c c c c |>{\columncolor{gray!10}}c }
    \toprule
    \textbf{Method} & \# \textbf{Params (\%)} & \textbf{BoolQ} & \textbf{PIQA} & \textbf{SIQA} & \textbf{HellaSwag} & \textbf{WinoGrande} & \textbf{ARC-e} & \textbf{ARC-c} & \textbf{OBQA} & \textbf{Avg.} \\
    \toprule
    ChatGPT & - & 73.1 & 85.4 & 68.5 & 78.5 & 66.1 & 89.8 & 79.9 & 74.8 & 77.0 \\ \midrule
    
    AdaLoRA$_{r=32}$ & 0.83 & 69.1 & 82.2 & 77.2 & 78.3 & 78.2 & 79.7 & 61.9 & 77.2 & 75.5 \\  \rowcolor{gray!20}

    AdaLoRA-Dash & 0.83 & 69.2 & 82.3 & 77.5 & 78.4 & 77.1 & 80.1 & 63.4 & 79.6 & \textbf{76.0} \\
    
    \midrule
    
    FLoRA$_{r=32}$ & 0.83 & 66.4 & 81.3 & 77.1 & 75.6 & 77.1 & 77.2 & 62.4 & 77.6 & 74.3  \\  \rowcolor{gray!20}
    
    FLoRA-Dash & 0.83 & 69.8 & 81.9 & 78.0 & 83.3 & 79.6 & 79.1 & 62.7 & 79.4 & \textbf{76.7} \\ 
    
    \bottomrule
    \end{tabular}}
    \label{tab: flora and adalora of commonsense}
\end{table*}

\subsection{Further Discussion on Directions for Dash}

We here further discuss the impact of selecting different dash directions. The effectiveness of choosing top, bottom, or random directions has been a topic frequently mentioned in various studies. However, we find differing conclusions in different works, which complicates the understanding. For instance, \cite{meng2024pissa} and \cite{zhang2024spectral} suggest that selecting top directions yields the best results, while \cite{hameed2024rosa} advocates for random selections, and reference \cite{wang2024milora} supports choosing the bottom. This inconsistency warrants further exploration.

Fine-tuning adapts a model’s pre-trained knowledge to task-specific details. TSDs correspond to the directions with the most significant weight changes during adaptation. Intuitively, these should be the most impactful for fine-tuning.
TSDs typically align with directions of moderate singular values, suggesting that dashing in bottom directions might overlap with TSDs but yield worse results. 
Top directions, containing more generalized pre-training knowledge, could degrade performance when fine-tuning for task-specific details. Random direction selection introduces unpredictability, occasionally matching TSDs but generally performing worse than bottom selections.

In our empirical tests, the results largely align with our predictions. However, there are some exceptions where the effectiveness of top directions surpasses random or bottom, or where the bottom directions approach or even slightly exceed the performance of the TSDs. Please look for a counterexample in the Fig. \ref{fig: ablation more}(a).
We consider these variations to be within the expected range, acknowledging that it’s unrealistic to expect uniform trends across all tasks and models. The trends we have identified represent a more general, broadly applicable pattern.

In addition to the standard experimental setup, we conducted two additional experiments: adjusting the coordinate value changes of all directions or both top and bottom directions collectively in Fig. \ref{fig: ablation more}. Neither of these configurations resulted in improved outcomes. This aligns with our analysis in Sec. \ref{sec: fur dash num and length}, which suggested that the inclusion of irrelevant directions could degrade training performance due to increased noise in the model updates.
\section{Conclusion}\label{sec conclusion}

In this paper, we revisit LoRA’s exploration of task-specific directions (TSDs), highlighting its inadequacies in understanding TSDs. We then provide a precise definition of TSD and delve into their properties to better understand their role in fine-tuning large language models. Building on this foundational knowledge, we introduce LoRA-Dash, LoRA-Init, and LoRA-TSD, designed to fully unleash the potential of TSDs. Through comprehensive experiments and in-depth analysis, we not only demonstrate the significant advantages of our methods over conventional methods, but also highlight the crucial importance of TSDs in achieving superior task-specific performance. 
By pushing the boundaries of parameter-efficient fine-tuning, we aim to inspire continued research and development in this vibrant area, transforming practices across diverse applications in natural language processing and beyond.
 
\bibliographystyle{IEEEtran}
\bibliography{main}

\begin{IEEEbiography}
[{\includegraphics[width=1in,height=1.25in,clip,keepaspectratio]{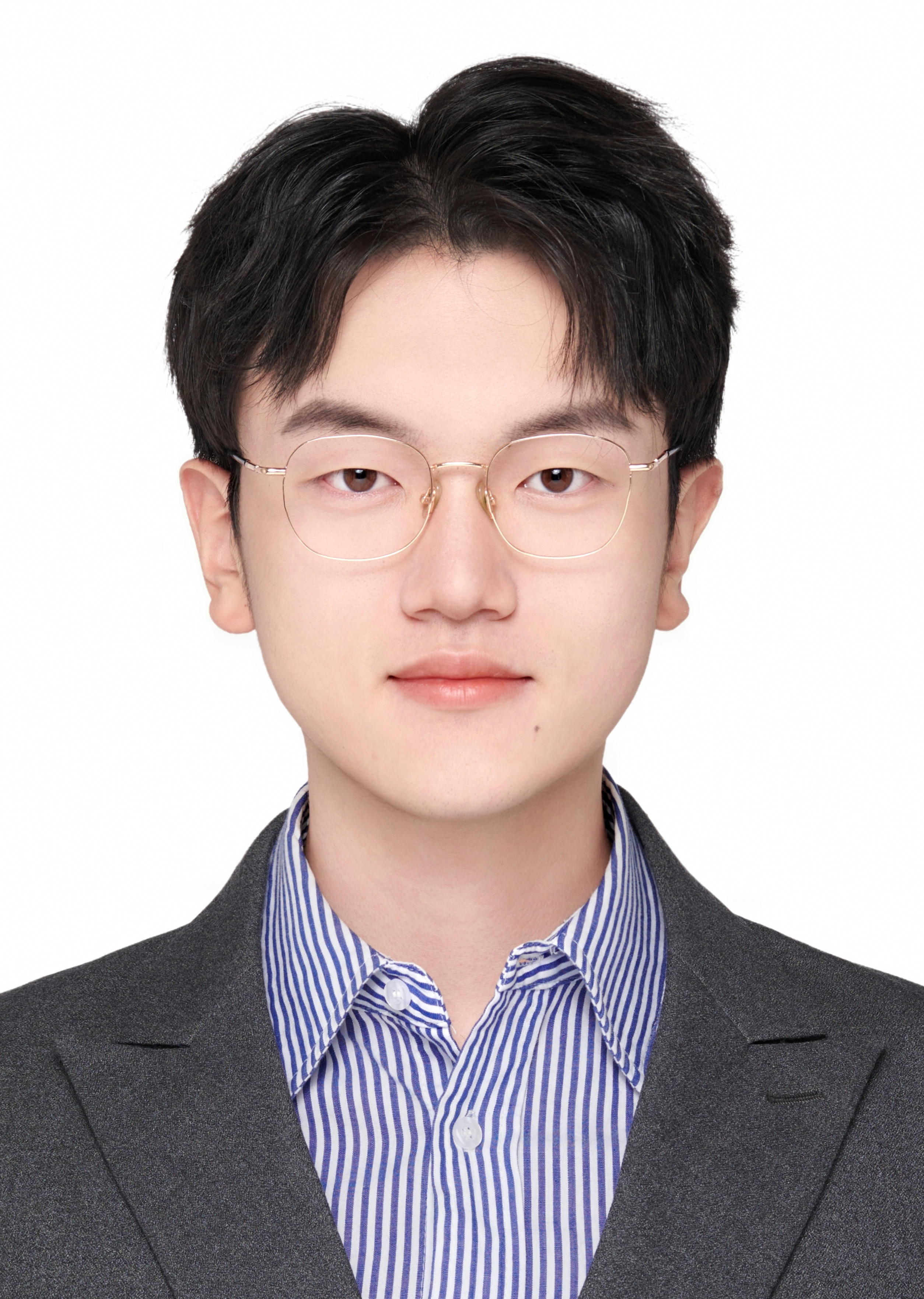}}]
{Chongjie Si} received the B.S. degree in artificial intelligence from Chien-Shiung Wu College, Southeast University, Nanjing, China, in 2022. 
He is currently working toward the PhD degree in the Artificial Intelligence Institute, Shanghai Jiao Tong University, Shanghai, China. His current research interests lie in efficient LLM training, including efficient pre-training and parameter efficient fine-tuning.
\end{IEEEbiography}

\begin{IEEEbiography}
[{\includegraphics[width=1in,height=1.25in,clip,keepaspectratio]{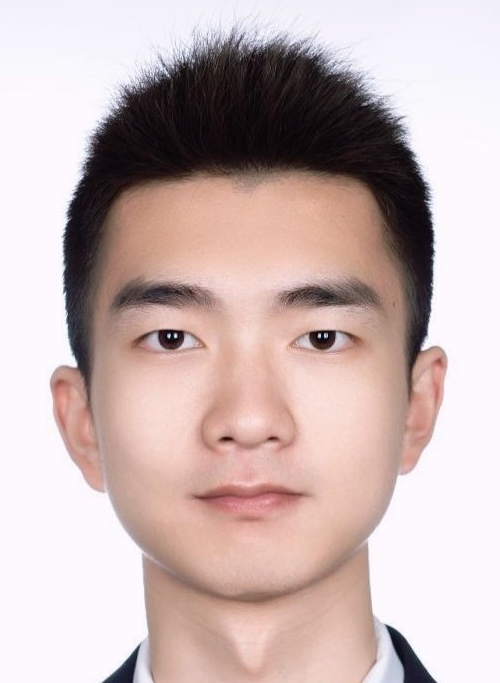}}]
{ Zhiyi Shi} received the Bachelor of Science degree from Chien-Shiung Wu College, Southeast University, Nanjing, China, in 2022. He then pursued his Master of Science degree at Carnegie Mellon University, Pittsburgh, PA, USA, which he completed in 2024. Currently, he is working as a Research Assistant at the Visual Computing Group Lab at Harvard University. His research interests include multimodal learning, large language models, and medical AI.
\end{IEEEbiography}

\begin{IEEEbiography}
[{\includegraphics[width=1in,height=1.25in,clip,keepaspectratio]{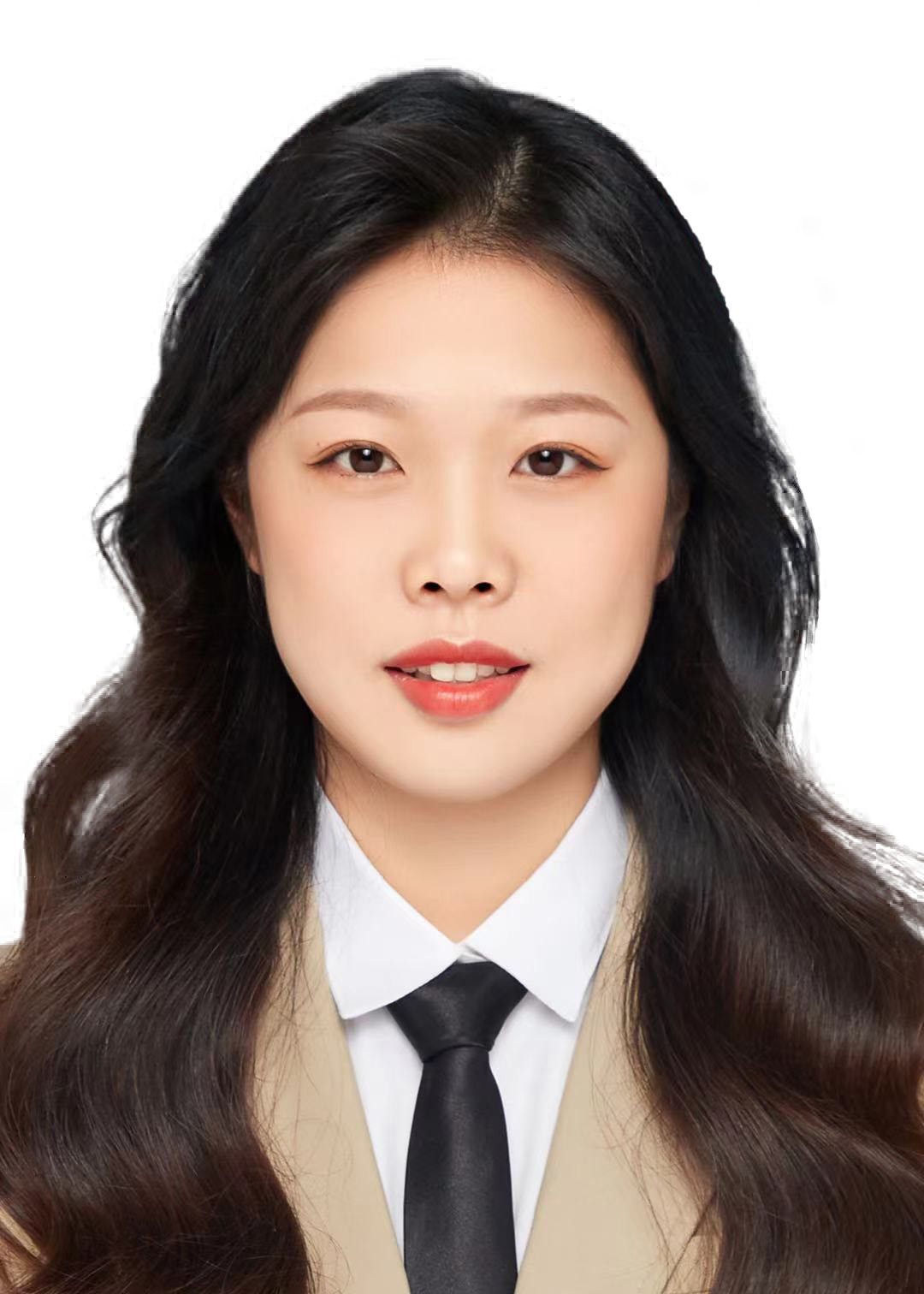}}]
{Zhang Shifan} received the bachelor’s degree in Electrical Engineering from Huazhong University of Science and Technology, Wuhan, China, in 2022. She is currently working toward the Ph.D. degree in Computer Science with the School of Electronic Information and Electrical Engineering, Shanghai Jiao Tong University, Shanghai, China. Her research interests include autonomous driving and computer vision.
\end{IEEEbiography}

\begin{IEEEbiography}
[{\includegraphics[width=1in,height=1.25in,clip,keepaspectratio]{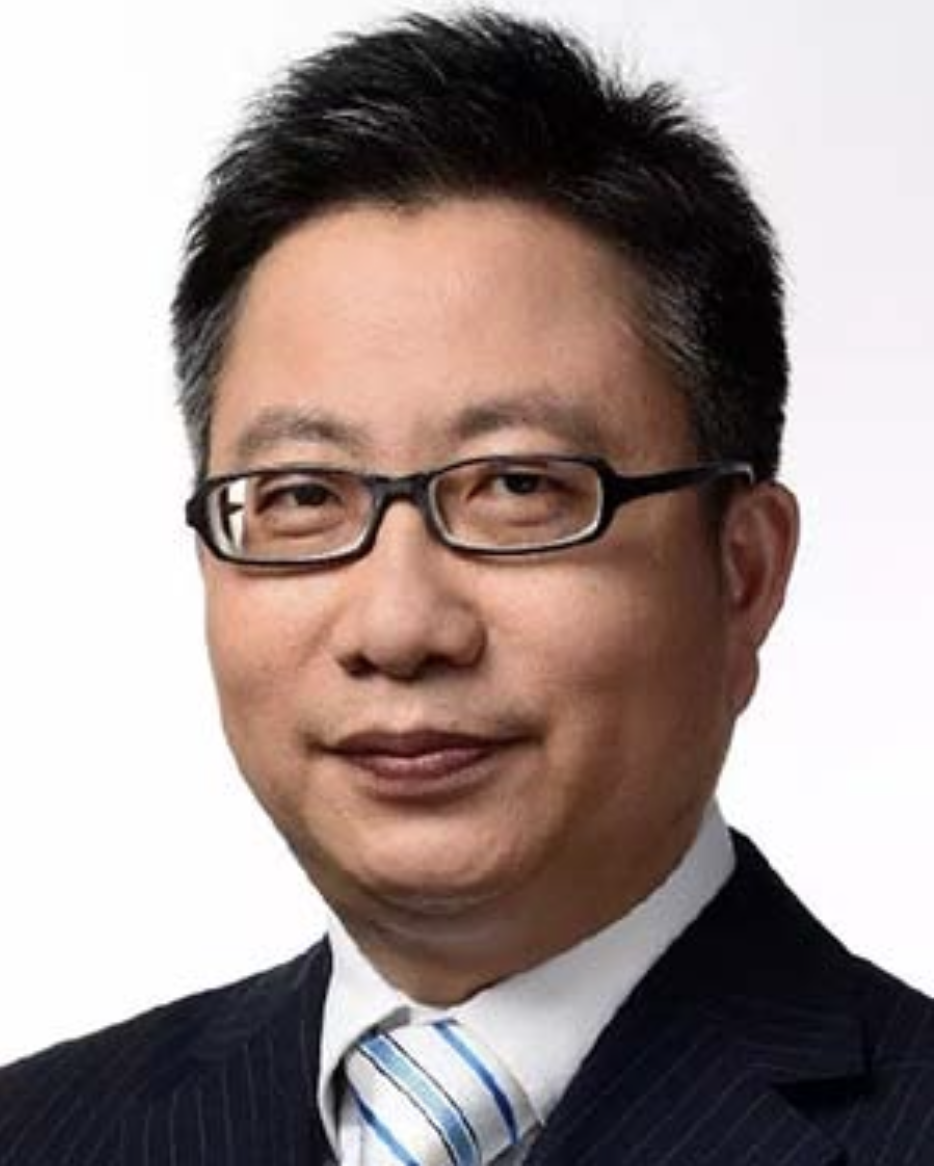}}]
{Xiaokang Yang} (Fellow, IEEE) received the BS degree from Xiamen University, Xiamen, China, in 1994, the MS degree from the Chinese Academy of Sciences, Shanghai, China, in 1997, and the PhD degree from Shanghai Jiao Tong University, Shanghai, in 2000. From 2000 to 2002, he worked as a research fellow at the Centre for Signal Processing, Nanyang Technological University, Singapore. From 2002 to 2004, he was a research scientist at the Institute for Infocomm Research (I2R), Singapore. From 2007 to 2008, he visited the Institute for Computer Science, University of Freiburg, Germany, as an Alexander von Humboldt research fellow. He is currently a distinguished professor with the School of Electronic Information and Electrical Engineering, Shanghai Jiao Tong University. His current research interests include image processing and communication, computer vision, and machine learning. He is an associate editor of IEEE Transactions on Multimedia and a senior associate editor of the IEEE Signal Processing Letters.
\end{IEEEbiography}

\begin{IEEEbiography}
[{\includegraphics[width=1in,height=1.25in,clip,keepaspectratio]{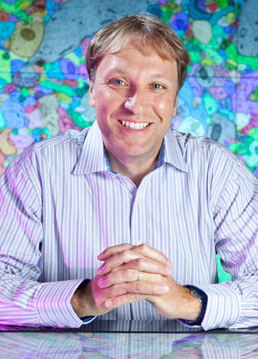}}]
{Hanspeter Pfister} (Fellow, IEEE) received an M.Sc. in electrical engineering from ETH Zurich, Switzerland, in 1991, and a Ph.D. in computer science from the State University of New York at Stony Brook in 1996. He is currently the An Wang Professor of Computer Science at the Harvard John A. Paulson School of Engineering and Applied Sciences and an affiliate faculty member of the Center for Brain Science and the D3 Institute at Harvard Business School. His research in visual computing lies at the intersection of visualization, computer graphics, and computer vision. It spans various topics, including biomedical visualization, image and video analysis, interpretable machine learning, and data science. From 2013 to 2017, he served as Director of the Institute for Applied Computational Science. Before joining Harvard, he worked for over a decade at Mitsubishi Electric Research Laboratories as Associate Director and Senior Research Scientist. He was the chief architect of VolumePro, Mitsubishi Electric's award-winning real-time volume rendering graphics card, for which he received the Mitsubishi Electric President's Award in 2000. He was elected an ACM Fellow in 2019 and an IEEE Fellow in 2023.
\end{IEEEbiography}

\begin{IEEEbiography}
[{\includegraphics[width=1in,height=1.25in,clip,keepaspectratio]{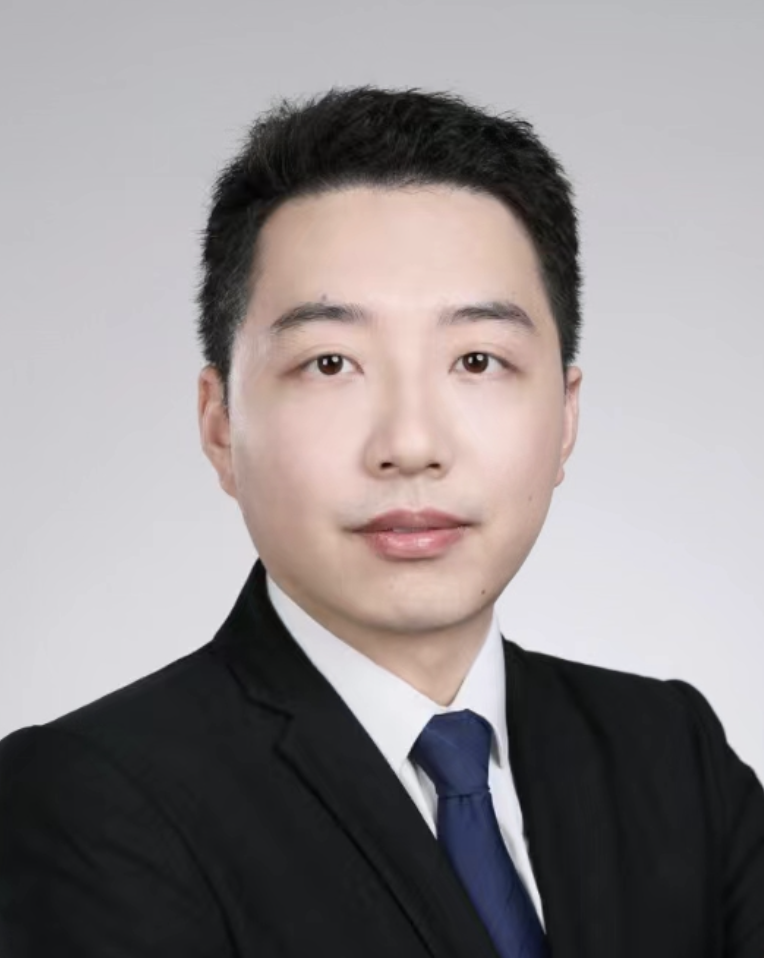}}]
{Wei Shen} is a professor at the Artificial Intelligence Institute, Shanghai Jiao Tong University. He was an Assistant Research Professor at the Department of Computer Science, Johns Hopkins University. He has over 80 peer-reviewed publications in computer vision and machine learning related areas, including IEEE Trans. PAMI, IEEE Trans. Image Processing, IEEE Trans. Medical Imaging, NeurIPS, ICML, ICCV, CVPR, ECCV and MICCAI. He is an Area Chair for ICCV 2025, NeurIPS 2023/2024, CVPR 2022/2023, ACCV 2022 and WACV 2024. He is an Associate Editor for Pattern Recognition. He receives the MICCAI Young Scientist Award and the NSFC Excellent Young Scientists Fund in 2023.
\end{IEEEbiography}

\newpage


\section{Details on Experiments}

\subsection{Details on Fully Fine-tuning LLaMA in Section \ref{sec: pre property}}\label{sec: supp detail ft LLaMA}

To further explore the properties of TSDs, we fully fine-tune LLaMA-7B \cite{touvron2023llama} on commonsense reasoning tasks. For the commonsense reasoning tasks, which consist of 8 distinct sub-tasks, each defined with specific training and testing sets, we have adopted the approach in \cite{hu2023llm}. Following this method, we amalgamated the training datasets from all sub-tasks to construct a comprehensive final training dataset. More details can be found in Sec. \ref{sec: supp detail cr task}. The experiment was conducted on 8 A100 GPUs.

The change rates in Fig. \ref{fig:change W star} are derived from the weights obtained from the self-attention value projection in the 16th layer. However, our validation confirms that the results and conclusions drawn from the diagrams are consistent, regardless of which layer or specific module within the attention mechanism is analyzed. For the original change rate $\delta$ for a core basis, it is scaled as $\ln(\delta+1)/3$.

\begin{table*}[!ht]

\centering
\renewcommand\arraystretch{1.1}
\setlength{\tabcolsep}{5.4mm}
\caption{Hyper-parameter settings of our methods on commonsense reasoning task.}
\resizebox{\textwidth}{!}{
\begin{tabular}{c | c c c c c | c c | c c} 

\toprule

\textbf{Hyper-parameters} & \multicolumn{5}{c}{LLaMA-7B} & \multicolumn{2}{c}{LLaMA2-7B} & \multicolumn{2}{c}{LLaMA3-8B}\\ 

\midrule

Rank $r$ & 4 & 8 & 16 & 32 & 64 & 16 & 32 & 16 & 32 \\ \midrule

$\alpha$ & 8 & 16 & 32& 64 & 128 & 32 & 64 & 32 & 64 \\ \midrule

LR & 5e-4 & 4e-4 & 5e-4 & 1e-4 & 0.9e-4 & 2e-4 & 1e-4 & 2e-4 & 0.8e-4 \\ \midrule

LR Scheduler & \multicolumn{9}{c}{Linear} \\ \midrule

Dropout & \multicolumn{9}{c}{0.05} \\ \midrule

Optimizer & \multicolumn{9}{c}{AdamW} \\ \midrule

Batch size & \multicolumn{9}{c}{16} \\ \midrule

Warmup Steps & \multicolumn{9}{c}{100} \\ \midrule

Epochs & \multicolumn{9}{c}{3} \\ \midrule

Where & \multicolumn{9}{c}{Q, K, V, Up, Down} \\

\bottomrule

\end{tabular}}
\label{tab: cr detail}
\end{table*}

\subsection{Details on Commonsense Reasoning Task}\label{sec: supp detail cr task}

The commonsense reasoning benchmarks consist of 8 distinct sub-tasks, each with its designated dataset, i.e., BoolQ \cite{clark2019boolq}, PIQA \cite{bisk2020piqa}, SIQA \cite{sap2019socialiqa}, HellaS. \cite{zellers2019hellaswag}, WinoG. \cite{sakaguchi2021winogrande}, ARC-e/ARC-c \cite{clark2018thinkarce}, OBQA \cite{mihaylov2018canobqa}. Adhering to the protocol outlined by \cite{hu2023llm}, we merge the training datasets from all tasks to form a comprehensive training dataset (Commonsense170K dataset), subsequently performing evaluations against each individual task’s testing set. 

We fine-tune LLaMA-7B \cite{touvron2023llama}, LLaMA2-7B \cite{touvron2023llama2} and LLaMA3-8B \cite{llama3modelcard} on this task. Additionally, we integrate results from ChatGPT’s implementation with the gpt-3.5-turbo API, particularly focusing on zero-shot Chain of Thought approaches \cite{wei2022chain}.

To ensure equitable comparisons, we standardize the training setup across all methods. 
For LoRA-Dash, LoRA-Init, and LoRA-TSD, we adopt the same LoRA configuration (rank, scaling factor, dropout, and frozen backbone) and only vary the learning rate to optimize performance on the validation set. 
Concretely, all compared methods are trained with the Adam optimizer for 3 epochs using a linear learning-rate schedule, without early stopping; models are evaluated once after training with a shared evaluation script.
We use the same random seed (42) for all runs to control stochasticity, and the training budget (number of epochs/updates over the same dataset) is therefore matched across methods to the greatest extent possible. 
The full hyper-parameter settings of our methods are listed in Table \ref{tab: cr detail}. 
For the vanilla LoRA baseline (LLaMA-7B / LLaMA2-7B / LLaMA3-8B, $r=32$), we directly cite the results reported in \cite{hu2023llm, liu2024dora} to avoid potential discrepancies introduced by re-implementation.
For other LoRA ranks, we only change the learning rate of $r=4/8$ at $2e-4$, and keep the others $3e-4$.


\begin{table*}[!ht]
    \centering
    \caption{Details of GLUE dataset.}
    \setlength{\tabcolsep}{5.4mm}
    \resizebox{\textwidth}{!}{
    \begin{tabular}{l | l  c  c  c  c  c}
    \toprule
         Dataset & Task & \# Train & \# Dev & \# Test & \# Label & Metrics \\ \midrule
         \multicolumn{7}{c}{Single-Sentence Classification} \\ \hline
         
         CoLA & Acceptability & 8.5k & 1k & 1k & 2 & Matthews corr \\ \midrule
         
         SST-2 & Sentiment & 67k & 872 & 1.8k & 2 & Accuracy \\ \midrule
         
         \multicolumn{7}{c}{Similarity and Paraphrase} \\ \midrule

         MRPC & Paraphrase & 3.7k & 408 & 1.7k & 2 & Accuracy / F1 \\ \midrule

         QQP & Paraphrase & 364k & 40k & 391k & 2 & Accuracy / F1 \\ \midrule
         
         STS-B & Similarity & 7k & 1.5k & 1.4k & 1 & Pearson/ Spearman Corr \\  \midrule

        \multicolumn{7}{c}{Natural Language Inference} \\ \midrule
          
         MNLI & NLI & 393k & 20k & 20k & 3 & Accuracy \\ \midrule
         
         QNLI & QA/NLI & 108k & 5.7k & 5.7k & 2 & Accuracy \\ \midrule

         RTE & NLI & 2.5k & 276 & 3k & 2 & Accuracy \\
        
         \bottomrule
    \end{tabular}}
    \label{tab: glue dataset}
\end{table*}

\begin{table*}
\centering
\caption{Hyper-parameter settings of our methods on NLU task.}
\setlength{\tabcolsep}{4mm}
\resizebox{\textwidth}{!}{
\begin{tabular}{c | c c c c c c c c} 

\toprule

Hyper-parameter & MNLI & SST-2 & CoLA & QQP & QNLI & RTE & MRPC & STS-B\\ 

\midrule

Optimizer & \multicolumn{8}{c}{AdamW} \\ \midrule

Warmup Ratio & \multicolumn{8}{c}{0.1} \\ \midrule

LR schedule & \multicolumn{8}{c}{Linear} \\  \midrule

Rank $r$ & \multicolumn{8}{c}{2 \& 8}\\ \midrule

LoRA alpha & \multicolumn{8}{c}{4 \& 16} \\ \midrule

Max Seq. Len. & 256 & 128 & 64 & 320 & 512 & 320 & 320 & 128 \\ \midrule

Batch Size & 32 & 32 & 32 & 32 & 32 & 32 & 32 & 32 \\ \midrule

Learning Rate & 5e-4 & 8e-4 & 8e-4 & 1e-3 & 5e-4 & 1.2e-3 & 1e-3 & 5e-4 \\ \midrule

Epochs & 12 & 24 & 25 & 5 & 5 & 50 & 30 & 25  \\ 

\bottomrule

\end{tabular}}
\label{tab: nlu detail}
\end{table*}

\begin{table*}[!ht]
 \renewcommand\arraystretch{1.1}
 \setlength{\tabcolsep}{1mm}
    \centering
    \caption{Precision/recall ($\times100\%$) across all query, key, and value layers (total 96 layers) for each 1000 training steps with mean and standard deviation.}
    \resizebox{\textwidth}{!}{
    \begin{tabular}{ c c c c c c c c c c c}
    \toprule
     \textbf{Steps} & \textbf{1000} & \textbf{2000} & \textbf{3000} & \textbf{4000} & \textbf{5000} & \textbf{6000} & \textbf{7000} & \textbf{8000} & \textbf{9000} & \textbf{10000}  \\
     \toprule
     
    \textbf{Precision} & 78.9 $\pm$ 12.1 & 79.7 $\pm$ 12.2 & 77.6 $\pm$ 12.7 & 77.5 $\pm$ 12.9 & 78.5 $\pm$ 11.9 & 80.5 $\pm$ 12.1 & 79.0 $\pm$ 12.3 & 79.9 $\pm$ 12.8 & 79.0 $\pm$ 12.2 & 78.9 $\pm$ 13.2 \\ 
    
    \textbf{Recall} & 75.8 $\pm$ 18.7 & 75.5 $\pm$ 17.8 & 73.2 $\pm$ 18.9 & 74.0 $\pm$ 21.7 & 75.8 $\pm$ 19.0 & 79.2 $\pm$ 16.5 & 76.8 $\pm$ 19.9 & 78.4 $\pm$ 19.1 & 77.3 $\pm$ 19.6 & 77.6 $\pm$ 19.0  \\ 
    
    \midrule

     \textbf{Steps} & \textbf{11000} & \textbf{12000} & \textbf{13000} & \textbf{14000} & \textbf{15000} & \textbf{16000} & \textbf{17000} & \textbf{18000} & \textbf{19000} & \textbf{20000}  \\
     \toprule

     \textbf{Precision} & 79.0 $\pm$ 11.4 & 78.8 $\pm$ 11.6 & 78.8 $\pm$ 12.3 & 80.3 $\pm$ 12.7 & 81.3 $\pm$ 12.7 & 81.0 $\pm$ 13.2 & 81.0 $\pm$ 12.2 & 80.7 $\pm$ 12.2 & 81.4 $\pm$ 12.6 & 81.3 $\pm$ 12.3 \\

    \textbf{Recall} & 78.4 $\pm$ 19.1 & 77.9 $\pm$ 17.3 & 78.4 $\pm$ 18.7 & 76.6 $\pm$ 21.0 & 78.4 $\pm$ 18.7 & 78.4 $\pm$ 19.4 & 78.1 $\pm$ 20.3 & 78.1 $\pm$ 19.6 & 77.1 $\pm$ 20.4 & 76.6 $\pm$ 18.8 \\ \midrule

    \textbf{Steps} & \textbf{21000} & \textbf{22000} & \textbf{23000} & \textbf{24000} & \textbf{25000} & \textbf{26000} & \textbf{27000} & \textbf{28000} & \textbf{29000} & \textbf{30000}  \\

    \toprule

    \textbf{Precision} & 80.1 $\pm$ 14.1 & 79.4 $\pm$ 13.1 & 80.9 $\pm$ 12.9 & 79.7 $\pm$ 12.3 & 80.2 $\pm$ 11.7 & 79.9 $\pm$ 12.2 & 79.3 $\pm$ 12.7 & 78.8 $\pm$ 13.3 & 79.3 $\pm$ 12.5 & 79.7 $\pm$ 12.3 \\

    \textbf{Recall} & 74.7 $\pm$ 22.2 & 76.8 $\pm$ 19.6 & 75.8 $\pm$ 21.0 & 76.8 $\pm$ 19.9 & 77.3 $\pm$ 18.8 & 78.4 $\pm$ 18.4 & 77.1 $\pm$ 18.0 & 75.3 $\pm$ 19.3 & 76.0 $\pm$ 18.8 & 76.3 $\pm$ 19.3 \\
    
    \bottomrule
    \end{tabular}}
    \label{tab:results across layer}
\end{table*}

\begin{table*}[!ht]
 \renewcommand\arraystretch{1.1}
 \setlength{\tabcolsep}{1mm}
    \centering
    \caption{Precision/recall ($\times100\%$) across all steps for each query (Q), key (K), and value (V) layer with mean and standard deviation.}
    \resizebox{\textwidth}{!}{
    \begin{tabular}{ c c c c c c c c c c }
    \toprule
     \textbf{(Layer, Module)} & \textbf{(0, Q)} & \textbf{(0, K)} & \textbf{(0, V)} & \textbf{(1, Q)} & \textbf{(1, K)} & \textbf{(1, V)} & \textbf{(2, Q)} & \textbf{(2, K)} & \textbf{(2, V)} \\
     \toprule

    \textbf{Precision} & 71.2 $\pm$ 10.0 & 69.4 $\pm$ 11.0 & 74.7 $\pm$ 8.4 & 83.0 $\pm$ 7.3 & 84.8 $\pm$ 9.2 & 82.7 $\pm$ 10.5 & 86.0 $\pm$ 4.3 & 94.4 $\pm$ 10.6 & 75.4 $\pm$ 13.5  \\ 
    
    \textbf{Recall} & 64.2 $\pm$ 14.5 & 65.3 $\pm$ 18.4 & 91.3 $\pm$ 13.3 & 82.8 $\pm$ 12.7 & 88.7 $\pm$ 12.5 & 67.9 $\pm$ 13.0 & 58.9 $\pm$ 12.5 & 78.3 $\pm$ 17.3 & 72.5 $\pm$ 23.1 \\  \midrule

    \textbf{(Layer, Module)} & \textbf{(3, Q)} & \textbf{(3, K)} & \textbf{(3, V)} & \textbf{(4, Q)} & \textbf{(4, K)} & \textbf{(4, V)} & \textbf{(5, Q)} & \textbf{(5, K)} & \textbf{(5, V)} \\
     \toprule

    \textbf{Precision} & 82.1 $\pm$ 9.0 & 83.1 $\pm$ 8.4 & 90.9 $\pm$ 10.0 & 73.4 $\pm$ 14.3 & 61.9 $\pm$ 14.4 & 73.1 $\pm$ 9.6 & 71.7 $\pm$ 7.9 & 88.2 $\pm$ 8.3 & 80.9 $\pm$ 7.0 \\ 
    
    \textbf{Recall} & 73.9 $\pm$ 12.5 & 92.1 $\pm$ 13.8 & 89.9 $\pm$ 15.0 & 74.4 $\pm$ 22.5 & 43.5 $\pm$ 16.7 & 63.3 $\pm$ 24.6 & 73.0 $\pm$ 14.7 & 99.5 $\pm$ 3.7 & 96.2 $\pm$ 10.7 \\  \midrule
    
    \textbf{(Layer, Module)} & \textbf{(6, Q)} & \textbf{(6, K)} & \textbf{(6, V)} & \textbf{(7, Q)} & \textbf{(7, K)} & \textbf{(7, V)} & \textbf{(8, Q)} & \textbf{(8, K)} & \textbf{(8, V)} \\
     \toprule

    \textbf{Precision} & 78.8 $\pm$ 9.0 & 65.1 $\pm$ 8.0 & 90.1 $\pm$ 10.7 & 78.0 $\pm$ 9.8 & 72.9 $\pm$ 10.0 & 77.5 $\pm$ 10.5 & 86.4 $\pm$ 4.8 & 92.9 $\pm$ 8.8 & 71.3 $\pm$ 11.2 \\ 
    
    \textbf{Recall} & 83.0 $\pm$ 14.8 & 64.4 $\pm$ 16.8 & 90.0 $\pm$ 12.4 & 53.8 $\pm$ 18.5 & 60.2 $\pm$ 21.9 & 86.1 $\pm$ 18.2 & 77.0 $\pm$ 15.6 & 93.3 $\pm$ 12.3 & 68.4 $\pm$ 12.2 \\  \midrule

    \textbf{(Layer, Module)} & \textbf{(9, Q)} & \textbf{(9, K)} & \textbf{(9, V)} & \textbf{(10, Q)} & \textbf{(10, K)} & \textbf{(10, V)} & \textbf{(11, Q)} & \textbf{(11, K)} & \textbf{(11, V)} \\
     \toprule

    \textbf{Precision} & 90.8 $\pm$ 6.3 & 88.4 $\pm$ 7.3 & 74.9 $\pm$ 10.6 & 72.0 $\pm$ 10.6 & 85.9 $\pm$ 11.2 & 73.3 $\pm$ 7.0 & 85.1 $\pm$ 9.2 & 82.9 $\pm$ 7.6 & 84.9 $\pm$ 13.2 \\ 
    
    \textbf{Recall} & 90.4 $\pm$ 14.7 & 73.6 $\pm$ 6.7 & 56.2 $\pm$ 17.0 & 82.4 $\pm$ 13.2 & 72.8 $\pm$ 10.4 & 70.6 $\pm$ 13.6 & 80.6 $\pm$ 18.8 & 55.9 $\pm$ 13.1 & 88.3 $\pm$ 13.8 \\  \midrule

    \textbf{(Layer, Module)} & \textbf{(12, Q)} & \textbf{(12, K)} & \textbf{(12, V)} & \textbf{(13, Q)} & \textbf{(13, K)} & \textbf{(13, V)} & \textbf{(14, Q)} & \textbf{(14, K)} & \textbf{(14, V)} \\
     \toprule

    \textbf{Precision} & 87.7 $\pm$ 8.1 & 74.7 $\pm$ 9.5 & 62.5 $\pm$ 18.8 & 94.5 $\pm$ 7.5 & 70.5 $\pm$ 10.2 & 76.2 $\pm$ 9.4 & 72.4 $\pm$ 8.3 & 97.1 $\pm$ 5.6 & 71.6 $\pm$ 8.6 \\ 
    
    \textbf{Recall} & 79.2 $\pm$ 11.7 & 75.2 $\pm$ 11.9 & 72.3 $\pm$ 20.1 & 74.2 $\pm$ 19.2 & 73.4 $\pm$ 17.6 & 77.6 $\pm$ 18.0 & 78.8 $\pm$ 11.7 & 98.7 $\pm$ 5.8 & 66.0 $\pm$ 16.7  \\  \midrule

    \textbf{(Layer, Module)} & \textbf{(15, Q)} & \textbf{(15, K)} & \textbf{(15, V)} & \textbf{(16, Q)} & \textbf{(16, K)} & \textbf{(16, V)} & \textbf{(17, Q)} & \textbf{(17, K)} & \textbf{(17, V)} \\
     \toprule

    \textbf{Precision} & 64.1 $\pm$ 8.6 & 81.9 $\pm$ 14.6 & 81.6 $\pm$ 8.8 & 77.7 $\pm$ 9.4 & 83.6 $\pm$ 9.4 & 83.9 $\pm$ 10.0 & 80.3 $\pm$ 6.4 & 75.5 $\pm$ 8.7 & 82.9 $\pm$ 7.6 \\ 
    
    \textbf{Recall} & 71.3 $\pm$ 13.0 & 75.4 $\pm$ 11.1 & 92.0 $\pm$ 16.3 & 73.7 $\pm$ 10.0 & 67.1 $\pm$ 12.0 & 82.3 $\pm$ 17.5 & 90.8 $\pm$ 16.0 & 78.6 $\pm$ 11.2 & 84.3 $\pm$ 13.5 \\  \midrule
    
      \textbf{(Layer, Module)} & \textbf{(18, Q)} & \textbf{(18, K)} & \textbf{(18, V)} & \textbf{(19, Q)} & \textbf{(19, K)} & \textbf{(19, V)} & \textbf{(20, Q)} & \textbf{(20, K)} & \textbf{(20, V)} \\
     \toprule

    \textbf{Precision} & 73.1 $\pm$ 7.6 & 83.0 $\pm$ 7.9 & 92.4 $\pm$ 7.7 & 83.5 $\pm$ 8.5 & 79.3 $\pm$ 9.6 & 81.0 $\pm$ 8.6 & 91.5 $\pm$ 9.8 & 82.4 $\pm$ 9.7 & 89.8 $\pm$ 8.6 \\ 
    
    \textbf{Recall} & 67.3 $\pm$ 17.4 & 74.7 $\pm$ 7.4 & 78.3 $\pm$ 18.6 & 76.8 $\pm$ 11.7 & 58.2 $\pm$ 13.5 & 79.3 $\pm$ 17.0 & 69.0 $\pm$ 14.2 & 76.6 $\pm$ 15.8 & 89.7 $\pm$ 13.2  \\  \midrule
    
      \textbf{(Layer, Module)} & \textbf{(21, Q)} & \textbf{(21, K)} & \textbf{(21, V)} & \textbf{(22, Q)} & \textbf{(22, K)} & \textbf{(22, V)} & \textbf{(23, Q)} & \textbf{(23, K)} & \textbf{(23, V)} \\
     \toprule

    \textbf{Precision} & 87.0 $\pm$ 13.5 & 83.4 $\pm$ 8.6 & 73.5 $\pm$ 10.5 & 73.6 $\pm$ 8.6 & 94.0 $\pm$ 13.0 & 82.5 $\pm$ 10.3 & 75.4 $\pm$ 9.0 & 74.6 $\pm$ 11.0 & 80.4 $\pm$ 8.2 \\ 
    
    \textbf{Recall} & 79.2 $\pm$ 15.4 & 87.5 $\pm$ 13.7 & 45.4 $\pm$ 14.3 & 91.5 $\pm$ 13.7 & 96.1 $\pm$ 9.5 & 63.6 $\pm$ 12.5 & 81.3 $\pm$ 13.6 & 86.5 $\pm$ 14.0 & 58.7 $\pm$ 13.9    \\  \midrule
    
      \textbf{(Layer, Module)} & \textbf{(24, Q)} & \textbf{(24, K)} & \textbf{(24, V)} & \textbf{(25, Q)} & \textbf{(25, K)} & \textbf{(25, V)} & \textbf{(26, Q)} & \textbf{(26, K)} & \textbf{(26, V)} \\
     \toprule

    \textbf{Precision} & 82.5 $\pm$ 10.9 & 83.3 $\pm$ 12.0 & 86.8 $\pm$ 8.5 & 72.3 $\pm$ 7.5 & 84.2 $\pm$ 5.0 & 77.0 $\pm$ 8.0 & 87.2 $\pm$ 11.2 & 68.4 $\pm$ 10.1 & 67.3 $\pm$ 7.4 \\ 
    
    \textbf{Recall} & 66.7 $\pm$ 13.8 & 79.1 $\pm$ 12.0 & 87.5 $\pm$ 13.0 & 77.9 $\pm$ 9.6 & 91.5 $\pm$ 14.3 & 77.7 $\pm$ 13.9 & 70.6 $\pm$ 15.1 & 85.7 $\pm$ 15.2 & 67.2 $\pm$ 16.0    \\  \midrule
    
      \textbf{(Layer, Module)} & \textbf{(27, Q)} & \textbf{(27, K)} & \textbf{(27, V)} & \textbf{(28, Q)} & \textbf{(28, K)} & \textbf{(28, V)} & \textbf{(29, Q)} & \textbf{(29, K)} & \textbf{(29, V)} \\
     \toprule

    \textbf{Precision} & 74.3 $\pm$ 11.2 & 76.3 $\pm$ 7.4 & 74.8 $\pm$ 10.1 & 81.0 $\pm$ 10.2 & 75.6 $\pm$ 8.6 & 92.9 $\pm$ 13.6 & 87.7 $\pm$ 8.6 & 78.1 $\pm$ 6.9 & 83.4 $\pm$ 9.8 \\ 
    
    \textbf{Recall} & 76.6 $\pm$ 8.8 & 91.2 $\pm$ 13.6 & 91.4 $\pm$ 13.6 & 44.9 $\pm$ 18.0 & 71.9 $\pm$ 15.5 & 78.2 $\pm$ 16.9 & 86.8 $\pm$ 17.4 & 83.1 $\pm$ 17.8 & 86.8 $\pm$ 13.0 \\  \midrule
    
      \textbf{(Layer, Module)} & \textbf{(30, Q)} & \textbf{(30, K)} & \textbf{(30, V)} & \textbf{(31, Q)} & \textbf{(31, K)} & \textbf{(31, V)} \\
     \toprule

    \textbf{Precision} & 76.0 $\pm$ 8.6 & 79.3 $\pm$ 11.6 & 84.2 $\pm$ 7.8 & 65.8 $\pm$ 6.5 & 61.9 $\pm$ 11.6 & 74.5 $\pm$ 6.5  \\ 
    
    \textbf{Recall} & 90.0 $\pm$ 18.7 & 73.6 $\pm$ 16.8 & 91.7 $\pm$ 15.5 & 76.2 $\pm$ 14.5 & 55.0 $\pm$ 19.1 & 84.7 $\pm$ 12.5  \\  
    
    \bottomrule
    \end{tabular}}
    \label{tab:results across step}
\end{table*}

\begin{table*}[!ht]
 \renewcommand\arraystretch{1.1}
 \setlength{\tabcolsep}{1mm}
    \centering
    \caption{Precision/recall ($\times100\%$) across all query, key, and value layers (total 96 layers) for each 1000 training steps with mean and standard deviation on CoLA dataset.}
    \resizebox{\textwidth}{!}{
    \begin{tabular}{ c c c c c c c c c c c}
    \toprule
     \textbf{Steps} & \textbf{1000} & \textbf{2000} & \textbf{3000} & \textbf{4000} & \textbf{5000} & \textbf{6000} & \textbf{7000} & \textbf{8000} & \textbf{9000} & \textbf{10000}  \\
     \toprule
     
    \textbf{Precision} 
    & 78.8 $\pm$ 11.7 & 79.4 $\pm$ 11.7 & 77.7 $\pm$ 13.3 & 78.4 $\pm$ 13.7 & 78.4 $\pm$ 12.0 
    & 80.7 $\pm$ 12.8 & 78.2 $\pm$ 11.6 & 79.6 $\pm$ 13.7 & 78.2 $\pm$ 12.6 & 79.4 $\pm$ 12.7 \\ 
    
    \textbf{Recall} 
    & 76.1 $\pm$ 19.6 & 75.9 $\pm$ 17.3 & 74.1 $\pm$ 19.5 & 73.2 $\pm$ 20.9 & 75.6 $\pm$ 18.3 
    & 80.0 $\pm$ 16.3 & 76.1 $\pm$ 20.3 & 79.2 $\pm$ 19.4 & 76.6 $\pm$ 18.9 & 77.4 $\pm$ 18.4 \\ 
    
    \midrule

     \textbf{Steps} & \textbf{11000} & \textbf{12000} & \textbf{13000} & \textbf{14000} & \textbf{15000} & \textbf{16000} & \textbf{17000} & \textbf{18000} & \textbf{19000} & \textbf{20000}  \\
     \toprule

     \textbf{Precision} 
     & 79.1 $\pm$ 11.5 & 79.0 $\pm$ 11.2 & 78.2 $\pm$ 11.7 & 79.7 $\pm$ 12.2 & 81.0 $\pm$ 13.4 
     & 81.3 $\pm$ 12.8 & 81.7 $\pm$ 11.8 & 79.9 $\pm$ 12.7 & 80.6 $\pm$ 12.1 & 80.5 $\pm$ 11.6 \\ 

    \textbf{Recall} 
    & 77.9 $\pm$ 18.7 & 77.7 $\pm$ 17.1 & 79.2 $\pm$ 19.6 & 76.3 $\pm$ 21.4 & 78.9 $\pm$ 19.2 
    & 79.2 $\pm$ 19.7 & 78.7 $\pm$ 19.9 & 77.7 $\pm$ 20.2 & 76.4 $\pm$ 20.2 & 76.5 $\pm$ 19.2 \\ 
    
    \midrule

    \textbf{Steps} & \textbf{21000} & \textbf{22000} & \textbf{23000} & \textbf{24000} & \textbf{25000} & \textbf{26000} & \textbf{27000} & \textbf{28000} & \textbf{29000} & \textbf{30000}  \\
    \toprule

    \textbf{Precision} 
    & 80.8 $\pm$ 13.9 & 78.8 $\pm$ 13.8 & 80.2 $\pm$ 12.6 & 80.1 $\pm$ 12.4 & 80.4 $\pm$ 11.0 
    & 80.8 $\pm$ 13.0 & 79.6 $\pm$ 12.3 & 79.5 $\pm$ 13.0 & 78.5 $\pm$ 12.9 & 79.1 $\pm$ 11.4 \\

    \textbf{Recall} 
    & 74.2 $\pm$ 22.4 & 77.6 $\pm$ 19.9 & 76.2 $\pm$ 20.2 & 76.9 $\pm$ 19.2 & 76.5 $\pm$ 19.0 
    & 78.0 $\pm$ 17.6 & 78.0 $\pm$ 18.4 & 75.1 $\pm$ 20.1 & 76.2 $\pm$ 18.4 & 75.5 $\pm$ 19.9 \\
    
    \bottomrule
    \end{tabular}}
    \label{tab:results_across_layer_perturbed}
\end{table*}

\begin{table*}[!ht]
 \renewcommand\arraystretch{1.1}
 \setlength{\tabcolsep}{1mm}
    \centering
    \caption{Precision/recall ($\times100\%$) across all steps for each query (Q), key (K), and value (V) layer with mean and standard deviation on CoLA dataset.}
    \resizebox{\textwidth}{!}{
    \begin{tabular}{ c c c c c c c c c c }
    \toprule
     \textbf{(Layer, Module)} & \textbf{(0, Q)} & \textbf{(0, K)} & \textbf{(0, V)} & \textbf{(1, Q)} & \textbf{(1, K)} & \textbf{(1, V)} & \textbf{(2, Q)} & \textbf{(2, K)} & \textbf{(2, V)} \\
     \toprule

    \textbf{Precision} & 70.7 $\pm$ 9.2 & 69.9 $\pm$ 11.2 & 75.7 $\pm$ 9.2 & 82.7 $\pm$ 8.0 & 83.6 $\pm$ 10.1 & 83.1 $\pm$ 10.7 & 87.0 $\pm$ 3.3 & 94.0 $\pm$ 10.8 & 74.8 $\pm$ 13.7  \\ 
    
    \textbf{Recall} & 64.7 $\pm$ 14.8 & 66.0 $\pm$ 17.5 & 90.4 $\pm$ 14.1 & 81.9 $\pm$ 13.8 & 89.0 $\pm$ 12.7 & 69.0 $\pm$ 12.0 & 59.5 $\pm$ 13.3 & 77.7 $\pm$ 18.0 & 72.7 $\pm$ 23.6 \\  \midrule

    \textbf{(Layer, Module)} & \textbf{(3, Q)} & \textbf{(3, K)} & \textbf{(3, V)} & \textbf{(4, Q)} & \textbf{(4, K)} & \textbf{(4, V)} & \textbf{(5, Q)} & \textbf{(5, K)} & \textbf{(5, V)} \\
     \toprule

    \textbf{Precision} & 81.6 $\pm$ 9.2 & 83.6 $\pm$ 7.6 & 91.0 $\pm$ 10.4 & 74.0 $\pm$ 15.4 & 61.4 $\pm$ 15.1 & 73.4 $\pm$ 8.7 & 72.5 $\pm$ 8.5 & 87.6 $\pm$ 7.4 & 81.4 $\pm$ 6.5 \\ 
    
    \textbf{Recall} & 72.9 $\pm$ 13.2 & 92.0 $\pm$ 14.1 & 90.3 $\pm$ 14.5 & 75.6 $\pm$ 21.5 & 42.7 $\pm$ 16.5 & 62.9 $\pm$ 25.8 & 72.3 $\pm$ 14.6 & 99.1 $\pm$ 3.3 & 96.6 $\pm$ 11.0 \\  \midrule
    
    \textbf{(Layer, Module)} & \textbf{(6, Q)} & \textbf{(6, K)} & \textbf{(6, V)} & \textbf{(7, Q)} & \textbf{(7, K)} & \textbf{(7, V)} & \textbf{(8, Q)} & \textbf{(8, K)} & \textbf{(8, V)} \\
     \toprule

    \textbf{Precision} & 79.3 $\pm$ 8.6 & 64.8 $\pm$ 8.8 & 91.0 $\pm$ 9.7 & 77.2 $\pm$ 9.6 & 73.8 $\pm$ 10.2 & 78.3 $\pm$ 10.1 & 85.8 $\pm$ 4.6 & 93.3 $\pm$ 8.4 & 71.1 $\pm$ 10.8 \\ 
    
    \textbf{Recall} & 82.4 $\pm$ 15.8 & 65.2 $\pm$ 17.3 & 88.9 $\pm$ 11.4 & 52.6 $\pm$ 18.1 & 60.0 $\pm$ 23.0 & 85.7 $\pm$ 18.7 & 76.1 $\pm$ 14.6 & 93.9 $\pm$ 11.2 & 68.9 $\pm$ 12.4 \\  \midrule

    \textbf{(Layer, Module)} & \textbf{(9, Q)} & \textbf{(9, K)} & \textbf{(9, V)} & \textbf{(10, Q)} & \textbf{(10, K)} & \textbf{(10, V)} & \textbf{(11, Q)} & \textbf{(11, K)} & \textbf{(11, V)} \\
     \toprule

    \textbf{Precision} & 91.6 $\pm$ 5.9 & 89.0 $\pm$ 6.3 & 73.8 $\pm$ 9.5 & 72.4 $\pm$ 10.9 & 85.3 $\pm$ 10.5 & 74.1 $\pm$ 6.9 & 84.1 $\pm$ 9.4 & 84.1 $\pm$ 8.3 & 84.3 $\pm$ 12.1 \\ 
    
    \textbf{Recall} & 90.3 $\pm$ 15.5 & 73.7 $\pm$ 5.8 & 55.7 $\pm$ 16.7 & 81.8 $\pm$ 12.7 & 73.0 $\pm$ 10.1 & 71.7 $\pm$ 12.8 & 80.1 $\pm$ 18.7 & 55.4 $\pm$ 13.9 & 88.8 $\pm$ 14.6 \\  \midrule

    \textbf{(Layer, Module)} & \textbf{(12, Q)} & \textbf{(12, K)} & \textbf{(12, V)} & \textbf{(13, Q)} & \textbf{(13, K)} & \textbf{(13, V)} & \textbf{(14, Q)} & \textbf{(14, K)} & \textbf{(14, V)} \\
     \toprule

    \textbf{Precision} & 88.8 $\pm$ 8.5 & 74.9 $\pm$ 9.9 & 62.8 $\pm$ 18.0 & 95.0 $\pm$ 6.5 & 70.4 $\pm$ 9.8 & 76.9 $\pm$ 9.9 & 72.0 $\pm$ 7.9 & 98.3 $\pm$ 4.6 & 71.5 $\pm$ 8.5 \\ 
    
    \textbf{Recall} & 79.7 $\pm$ 11.9 & 75.1 $\pm$ 12.4 & 72.8 $\pm$ 20.6 & 73.3 $\pm$ 18.2 & 74.3 $\pm$ 18.6 & 77.3 $\pm$ 18.8 & 79.9 $\pm$ 12.3 & 99.5 $\pm$ 6.8 & 65.3 $\pm$ 16.8  \\  \midrule

    \textbf{(Layer, Module)} & \textbf{(15, Q)} & \textbf{(15, K)} & \textbf{(15, V)} & \textbf{(16, Q)} & \textbf{(16, K)} & \textbf{(16, V)} & \textbf{(17, Q)} & \textbf{(17, K)} & \textbf{(17, V)} \\
     \toprule

    \textbf{Precision} & 64.3 $\pm$ 8.1 & 81.7 $\pm$ 15.8 & 80.6 $\pm$ 7.8 & 76.7 $\pm$ 8.6 & 83.4 $\pm$ 9.7 & 83.3 $\pm$ 9.5 & 79.9 $\pm$ 5.5 & 76.3 $\pm$ 9.8 & 83.6 $\pm$ 6.5 \\ 
    
    \textbf{Recall} & 72.0 $\pm$ 13.6 & 75.1 $\pm$ 11.8 & 92.8 $\pm$ 16.0 & 74.0 $\pm$ 10.8 & 67.7 $\pm$ 10.8 & 83.1 $\pm$ 18.2 & 90.6 $\pm$ 16.6 & 79.0 $\pm$ 10.8 & 83.2 $\pm$ 13.9 \\  \midrule
    
      \textbf{(Layer, Module)} & \textbf{(18, Q)} & \textbf{(18, K)} & \textbf{(18, V)} & \textbf{(19, Q)} & \textbf{(19, K)} & \textbf{(19, V)} & \textbf{(20, Q)} & \textbf{(20, K)} & \textbf{(20, V)} \\
     \toprule

    \textbf{Precision} & 72.8 $\pm$ 6.8 & 82.5 $\pm$ 7.2 & 92.7 $\pm$ 7.0 & 83.3 $\pm$ 9.1 & 79.1 $\pm$ 9.5 & 80.3 $\pm$ 7.6 & 91.0 $\pm$ 9.3 & 83.1 $\pm$ 9.1 & 89.4 $\pm$ 9.2 \\ 
    
    \textbf{Recall} & 68.5 $\pm$ 16.7 & 75.9 $\pm$ 7.9 & 77.7 $\pm$ 18.7 & 76.1 $\pm$ 11.9 & 57.6 $\pm$ 13.7 & 79.4 $\pm$ 16.1 & 67.9 $\pm$ 13.4 & 75.9 $\pm$ 15.6 & 89.5 $\pm$ 13.4  \\  \midrule
    
      \textbf{(Layer, Module)} & \textbf{(21, Q)} & \textbf{(21, K)} & \textbf{(21, V)} & \textbf{(22, Q)} & \textbf{(22, K)} & \textbf{(22, V)} & \textbf{(23, Q)} & \textbf{(23, K)} & \textbf{(23, V)} \\
     \toprule

    \textbf{Precision} & 87.3 $\pm$ 13.0 & 84.5 $\pm$ 9.7 & 73.6 $\pm$ 10.0 & 73.1 $\pm$ 8.9 & 94.8 $\pm$ 12.6 & 83.7 $\pm$ 11.2 & 74.6 $\pm$ 9.7 & 75.7 $\pm$ 12.2 & 79.7 $\pm$ 8.0 \\ 
    
    \textbf{Recall} & 78.6 $\pm$ 16.6 & 88.5 $\pm$ 13.2 & 45.1 $\pm$ 14.5 & 91.2 $\pm$ 13.8 & 96.3 $\pm$ 10.4 & 64.3 $\pm$ 12.8 & 80.6 $\pm$ 12.5 & 85.4 $\pm$ 14.2 & 59.7 $\pm$ 14.0    \\  \midrule
    
      \textbf{(Layer, Module)} & \textbf{(24, Q)} & \textbf{(24, K)} & \textbf{(24, V)} & \textbf{(25, Q)} & \textbf{(25, K)} & \textbf{(25, V)} & \textbf{(26, Q)} & \textbf{(26, K)} & \textbf{(26, V)} \\
     \toprule

    \textbf{Precision} & 82.6 $\pm$ 11.6 & 82.7 $\pm$ 10.8 & 87.1 $\pm$ 7.6 & 71.9 $\pm$ 6.4 & 84.7 $\pm$ 6.2 & 76.3 $\pm$ 7.6 & 88.0 $\pm$ 11.0 & 68.0 $\pm$ 9.8 & 66.7 $\pm$ 6.4 \\ 
    
    \textbf{Recall} & 66.5 $\pm$ 13.7 & 79.8 $\pm$ 11.7 & 87.2 $\pm$ 12.2 & 77.6 $\pm$ 8.4 & 92.3 $\pm$ 15.3 & 78.4 $\pm$ 12.7 & 71.4 $\pm$ 14.3 & 85.1 $\pm$ 14.3 & 66.0 $\pm$ 16.7    \\  \midrule
    
      \textbf{(Layer, Module)} & \textbf{(27, Q)} & \textbf{(27, K)} & \textbf{(27, V)} & \textbf{(28, Q)} & \textbf{(28, K)} & \textbf{(28, V)} & \textbf{(29, Q)} & \textbf{(29, K)} & \textbf{(29, V)} \\
     \toprule

    \textbf{Precision} & 73.5 $\pm$ 10.6 & 76.6 $\pm$ 6.9 & 76.0 $\pm$ 9.5 & 81.4 $\pm$ 10.4 & 74.8 $\pm$ 9.7 & 92.0 $\pm$ 14.7 & 88.4 $\pm$ 7.6 & 78.6 $\pm$ 6.1 & 82.7 $\pm$ 9.3 \\ 
    
    \textbf{Recall} & 76.4 $\pm$ 9.8 & 92.1 $\pm$ 12.6 & 90.3 $\pm$ 14.1 & 44.1 $\pm$ 16.9 & 70.8 $\pm$ 14.4 & 77.3 $\pm$ 18.0 & 85.9 $\pm$ 17.7 & 84.0 $\pm$ 18.2 & 86.5 $\pm$ 13.8 \\  \midrule
    
      \textbf{(Layer, Module)} & \textbf{(30, Q)} & \textbf{(30, K)} & \textbf{(30, V)} & \textbf{(31, Q)} & \textbf{(31, K)} & \textbf{(31, V)} \\
     \toprule

    \textbf{Precision} & 76.4 $\pm$ 9.4 & 78.9 $\pm$ 12.6 & 83.3 $\pm$ 8.4 & 65.3 $\pm$ 6.6 & 60.7 $\pm$ 12.1 & 75.2 $\pm$ 7.4  \\ 
    
    \textbf{Recall} & 90.5 $\pm$ 19.3 & 72.7 $\pm$ 17.9 & 92.2 $\pm$ 15.7 & 77.0 $\pm$ 15.4 & 55.2 $\pm$ 19.7 & 85.8 $\pm$ 11.5  \\  
    
    \bottomrule
    \end{tabular}}
    \label{tab:results across step 2}
\end{table*}

\subsection{Details on Natural Language Understanding Task}\label{sec: supp detail nlu task}

For natural language understanding (NLU) task, we adopt the General Language Understanding Evaluation (GLUE) \cite{wang2018glue} benchmark, which is designed to test capabilities across various tasks. This benchmark consists of two single-sentence classification tasks, CoLA \cite{warstadt2019neural} and SST-2 \cite{socher2013recursive}, three similarity and paraphrase tasks, MRPC \cite{dolan2005automatically}, QQP \cite{wang2018glue}, and STS-B \cite{cer2017semeval}, and three natural language inference tasks, MNLI \cite{williams2017broad}, QNLI \cite{rajpurkar2016squad}, and RTE \cite{dagan2005pascal,bar2006second,giampiccolo2007third,bentivogli2009fifth}. The details of these datasets are shown in Table. \ref{tab: glue dataset}.

We fine-tune DeBERTaV3-base and DeBERTaV3-large \cite{he2021debertav3} models on this task. The hyper-parameter settings for this task is shown in Table. \ref{tab: nlu detail}.

\subsection{Details on Subject-driven Generation Task}\label{sec: supp detail sdg task}

In our experiment, we fine-tune the text-to-image diffusion models specifically tailored for subject-driven generation tasks, as outlined in recent research \cite{ruiz2023dreambooth}. The objective of this task is to generate images that adhere closely to prompts associated with a particular subject, defined by a few exemplar images. This involves initially fine-tuning a text-to-image model using image-text pairs where the text contains a unique identifier (e.g., ``A photo of a [V] cat''). Subsequent image generations are driven by new prompts incorporating this identifier, aiming to produce images aligned with the learned subject.

For this experiment, we use the SDXL5 model \cite{podell2023sdxl}, applying both LoRA and our methods for fine-tuning. The fine-tuning process is conducted with a learning rate of 1e-4 and a batch size of 4. We train the model over 500 steps on a single 80GB A100 GPU, taking approximately 23 minutes to complete. For the generation phase, we execute 50 inference steps for each given prompt to synthesize the final images, which takes approximately 7 seconds to complete.

We mainly adopt the official data of DreamBooth \cite{ruiz2023dreambooth} for diffusion.

\section{More Experiments}

\subsection{Discussion on LoRA's $\Delta\mathbf{W}$ as a Proxy}

We here present the statistical results of precision and recall metric shown in Fig. \ref{fig:accuracy TSD lora}, rank=32, on two tasks in Tables. \ref{tab:results across layer}-\ref{tab:results across step 2}. 

Note that a conceptual limitation of our formulation lies in the assumption of the existence of the optimal weight matrix $W^{\ast}$ that characterizes the true task-specific directions. 
In practice, however, $\mathbf{W}^{\ast}$ is inaccessible, and we instead approximate it with the empirical update $\Delta \mathbf{W}$ obtained via LoRA fine-tuning for PEFT. 
While this surrogate is empirically effective and produces stable statistical patterns across layers and training steps, the approximation error between $\Delta \mathbf{W}$ and the true $\mathbf{W}^{\ast} - \mathbf{W}$ remains analytically unquantified.
A more rigorous theoretical treatment would require establishing bounds on this approximation or characterizing conditions under which $\Delta \mathbf{W}$ reliably captures the intrinsic task-specific structure. 
We did not pursue such analysis in this work because deriving tight bounds typically demands strong assumptions on model architecture, optimization dynamics, and loss geometry—assumptions that are seldom satisfied in large-scale language models and would divert the focus away from our empirical investigation. 
Moreover, existing theoretical tools for analyzing low-rank updates in non-convex, over-parameterized networks remain limited, and a comprehensive treatment would constitute an independent line of research.
Given that our goal in this paper is to establish the empirical regularities, statistical behaviors, and practical utility of TSDs, we leave the development of a full theoretical characterization to future work.

\subsection{Training Costs}

In this subsection, we analyze the training efficiency of LoRA-TSD. 
As shown in Table \ref{tab:lora-tsd efficiency}, LoRA-TSD incurs moderately higher training time and memory usage compared to standard LoRA. 
This overhead mainly arises from the SVD factorization of the pre-trained weight matrix, and from the need to maintain the corresponding singular vectors. 
Although these matrices are fixed during training, LoRA-TSD still requires them to compute updates to the eight learnable singular values.
Consequently, training involves additional matrix operations over the eight singular vector basis, which explains the slight increase in computational and memory cost.

\begin{table}[ht]
    \centering
     \setlength{\tabcolsep}{2mm}
    \caption{Training time (hours) and memory consumption (per GPU) of LoRA-TSD and LoRA when training LLaMA-7B / LLaMA3-8B at $r=8$. LoRA-TSD (opt) refers to the version that we conduct efficiency optimization.}
    \begin{tabular}{c c | c c c}
    \toprule
        \multicolumn{2}{c|}{Method} & LoRA & LoRA-TSD & LoRA-TSD (opt) \\
        
        \midrule
        
       \multirow{2}{*}{LLaMA-7B} & Time & 4.0 & 5.9 & 5.5 \\
        & Memory & 37.3 & 45.3 & 39.8\\

        \midrule
        
        \multirow{2}{*}{LLaMA3-8B} & Time & 4.6 & 6.7 & 6.2 \\
        & Memory & 42.6 & 51.8 & 45.4\\
        
    \bottomrule
    \end{tabular}
    \label{tab:lora-tsd efficiency}
\end{table}

To further improve the training efficiency of LoRA-TSD, a practical optimization is to pre-compute the SVD of each target weight matrix offline, serialize the factors $\mathbf{U}$, $\mathbf{\Sigma}$, and $\pmb{V}$, and load them at runtime, thereby eliminating redundant SVD computation, reducing startup latency, and ensuring deterministic behavior across seeds.
For large models, memory usage can be reduced by lazily loading $\mathbf{U}$ and $\mathbf{V}$ from disk or CPU memory on demand, potentially combined with pinned memory and non-blocking transfers to overlap I/O with computation.
In multi-GPU or multi-node settings, precomputed SVDs can be shared via read-only memory-mapped files, avoiding duplicated storage across workers. Moreover, storing singular vectors in mixed precision (e.g., FP16/BF16) further decreases disk and VRAM cost with minimal impact on reconstruction quality.
By keeping $\mathbf{U}$ and $\mathbf{V}$ on CPU and placing only the small set of trainable singular values on GPU, LoRA-TSD becomes a lightweight GPU-resident optimization, improving both scalability and practical efficiency for large-scale model adaptation.

We adopt this pre-compute strategy for LoRA-TSD in our implementation. 
As shown in Table~\ref{tab:lora-tsd efficiency}, both training time and memory consumption of LoRA-TSD (opt) are further reduced, indicating that LoRA-TSD can be made practically efficient without sacrificing its performance benefits.
Importantly, this strategy only changes when the SVD is computed and how the singular vectors are loaded; it does not alter the training dynamics or the optimization process itself.
Consequently, the predictive performance of LoRA-TSD remains unchanged.
Given the substantial improvement in adaptation quality offered by our method, we believe that the minor increase in training cost—prior to applying this optimization—is negligible relative to the gains achieved.

\subsection{Experiments on Visual Instruction Tuning}

We also evaluate the effectiveness of LoRA-TSD on multi-modal models. For multi-modal tasks, we apply LoRA-TSD to fine-tune LLaVA-1.5-7B \cite{liu2024visual}, which consists of the Vicuna-1.5-7B language model \cite{peng2023instruction} and a CLIP ViT-L/336px vision encoder \cite{radford2021learning}, on visual instruction tuning. The fine-tuning is performed on seven standard vision–language benchmarks: VQA$^{v2}$ \cite{goyal2017making111}, GQA \cite{hudson2019gqa222}, VisWiz \cite{gurari2018vizwiz333}, SQA \cite{lu2022learn444}, VQA$^{\mathsf{T}}$ \cite{singh2019towards555}, POPE \cite{li2023evaluating666}, and MMBench \cite{liu2023mmbench777}.
The training hyper-parameters for the multi-modal experiments are summarized in Table \ref{tab:training5}. 
The corresponding experimental results are presented in Table \ref{tab:multi-modal}, where we can observe that LoRA-TSD still outperforms the baselines.

\begin{table}[ht]
    \centering
    \setlength{\tabcolsep}{5mm}
    \caption{Hyper-parameter configurations for fine-tuning LLaVA-1.5-7B with visual instruction tuning datasets.}
    \resizebox{0.9\linewidth}{!}{
    \begin{tabular}{c c}
    \toprule
       \textbf{Hyper-parameters}  & \textbf{Value}  \\ \midrule
        Rank $r$ &  128 \\
        $s$ & 8 \\
        $t$ & 100 \\
        Dropout & 0.05 \\
        Optimizer & AdamW \\
        LR & $2\times 10^{-4}$ \\
        LR Scheduler & Cosine decay \\
        Batch size & 16 \\
        Warmup ratio & 0.03 \\
        Epochs & 1 \\
        Where & Q,K,V,O,Up,Down,Gate \\
         \bottomrule
    \end{tabular}
    }
    \label{tab:training5}
\end{table}

\begin{table*}[ht]\footnotesize
\renewcommand\arraystretch{1}
\setlength{\tabcolsep}{3mm}
    \centering
    \caption{Results with LLaVA-1.5-7B \cite{liu2024visual} fine-tuned on visual instruction tuning.}
    \resizebox{\textwidth}{!}{
    \begin{tabular}{l | c| c c c c c c c |>{\columncolor{gray!10}}c}
    \toprule
       \multicolumn{1}{c|}{\textbf{Method}} & \textbf{Params} (\%) & \textbf{VQA$^{v2}$} & \textbf{GQA} & \textbf{VisWiz} & \textbf{SQA} & \textbf{VQA$^\mathsf{T}$} & \textbf{POPE} & \textbf{MMBench} & \textbf{Avg.}\\ \midrule
       
        Fully FT & 100 & 78.5 & 61.9 & 50.0 & 66.8 & 58.2 & 85.9 & 64.3 & 66.5 \\
        
        LoRA & 4.61 & 79.1 & 62.9 & 47.8 & 68.4 & 58.2 & 86.4 & 66.1 & 66.9\\
        
        \rowcolor{gray!20} 
        
        LoRA-TSD & 4.61 & 78.8 & 62.9 & 51.8 & 70.2 & 57.8 & 87.1 & 66.1 & \textbf{67.8}\\

         \bottomrule
    \end{tabular}
    }
    \label{tab:multi-modal}
\end{table*}

\subsection{Comparison with Other Methods}

\begin{table*}[!ht]
 \renewcommand\arraystretch{1}
 \setlength{\tabcolsep}{1.1mm}
    \centering
    \caption{Results on commonsense reasoning tasks of LoRA-TSD compared with other methods. For LoRA derivatives, we report the parameter gains over LoRA.}
    \resizebox{\textwidth}{!}{
    \begin{tabular}{ c c | c c c  c c c c c | c }
    \toprule
     \textbf{Method} & \textbf{Params} & \textbf{BoolQ} & \textbf{PIQA} & \textbf{SIQA} & \textbf{HellaS.} & \textbf{WinoG.} & \textbf{ARC-e} & \textbf{ARC-c} & \textbf{OBQA} & \textbf{Avg.} \\
    \toprule
    ChatGPT & - & 73.1 & 85.4 & 68.5 & 78.5 & 66.1 & 89.8 & 79.9 & 74.8 & 77.0 \\ \midrule
    
    \multicolumn{11}{c}{\textit{Fine-tuning LLaMA-7B}} \\ \midrule
    
     Fully FT & 100\% & 69.9 & 84.2 & 78.9 & 92.3 & 83.3 & 86.6 & 72.8 & 83.4 & 81.4 \\ \midrule

    Prefix & 0.11\% & 64.3 & 76.8 & 73.9 & 42.1 & 72.1 & 72.9 & 54.0 & 60.6 & 64.6 \\

    Series & 0.99\% & 63.0 & 79.2 & 76.3 & 67.9 & 75.7 & 74.5 & 57.1 & 72.4 & 70.8 \\

    Parallel & 3.54\% & 67.9 & 76.4 & 78.8 & 69.8 & 78.9 & 73.7 & 57.3 & 75.2 & 72.2 \\
    
    LoRA$_{r=4}$ & 0.10\% & 54.7 & 46.1 & 71.5 & 22.4 & 71.6 & 70.9 & 50.1 & 53.4 & 55.1 \\

    AdaLoRA$_{r=4}$ & + 0.6k & 66.1 & 78.1 & 74.3 & 34.0 & 74.4 & 76.7 & 57.5 & 71.2 & 66.5 \\

    FLoRA$_{r=4}$ & + 2.6k & \textbf{}67.2 & 78.0 & 72.9 & 65.4 & 73.8 & 73.8 & 55.3 & 71.8 & 69.8 \\

    DoRA$_{r=4}$ & + 877k & 51.3 & 42.2 & 77.8 & 25.4 & 78.8 & 78.7 & 62.5 & 78.6 & 61.9\\
    \rowcolor{gray!20} 

    LoRA-TSD$_{r=4}$ & + 1.3k & 66.7 & 80.3 & 78.3 & 81.9 & 78.8 & 79.2 & 64.7 & 79.1 & \textbf{76.1} \\

    \midrule
    
    LoRA$_{r=32}$ & 0.83\% & 68.9 & 80.7 & 77.4 & 78.1 & 78.8 & 77.8 & 61.3 & 74.8 & 74.7 \\

    AdaLoRA$_{r=32}$ & + 5.1k & 69.1 & 82.2 & 77.2 & 78.3 & 78.2 & 79.7 & 61.9 & 77.2 & 75.5 \\

    FLoRA$_{r=32}$ & + 164k & 66.4 & 81.3 & 77.1 & 75.6 & 77.1 & 77.2 & 62.4 & 77.6 & 74.3  \\ 

    DoRA$_{r=32}$ & + 877k & 69.7 & 83.4 & 78.6 & 87.2 & 81.0 & 81.9 & 66.2 & 79.2 & 78.4 \\
    
    \rowcolor{gray!20}
    
     LoRA-TSD$_{r=32}$ & + 1.3k & 69.8 & 82.6 & 78.7 & 85.1 & 81.6 & 81.4 & 66.9 & 81.0 & \textbf{78.4}  \\ \midrule

     \multicolumn{11}{c}{\textit{Fine-tuning LLaMA3-8B}} \\ \midrule

    LoRA$_{r=16}$ & 0.35\% & 72.3 & 86.7 & 79.3 & 93.5 & 84.8 & 87.7 & 75.7 & 82.8 & 82.8 \\

    AdaLoRA$_{r=16}$ & + 2.6k & 90.4 & 85.0 & 76.7 & 79.1 & 83.3 & 86.4 & 75.1 & 75.4 & 81.4 \\
    
    FLoRA$_{r=16}$ & + 41k & 90.2 & 84.2 & 79.9 & 79.3 & 85.1 & 86.7 & 74.8 & 93.9 & 84.2 \\

    \rowcolor{gray!20}

    LoRA-TSD$_{r=16}$ & + 1.3k & 7 74.8 &  88.3 & 80.9 & 95.5 & 86.9 & 90.4 & 78.3 & 85.0 & \textbf{85.0}  \\ \midrule

    LoRA$_{r=32}$ & 0.70\% & 70.8 & 85.2 & 79.9 & 91.7 & 84.3 & 84.2 & 71.2 & 79.0 & 80.8 \\

    PISSA$_{r=32}$ & + 0 & 67.1 & 81.1 & 77.2 & 83.6 & 78.9 & 77.7 & 63.2 & 74.6 & 75.4\\

    MiLoRA$_{r=32}$ & + 0 & 68.8 & 86.7 & 77.2 & 92.9 & 85.6 & 86.8 & 75.5 & 81.8 & 81.9 \\

    DoRA$_{r=32}$ & + 784k & 74.6 & 89.3 & 79.9 & 95.5 & 85.6 & 90.5 & 80.4 & 85.8 & 85.2 \\

    \rowcolor{gray!20}

    LoRA-TSD$_{r=32}$ & + 1.3k & 75.0 & 88.6 & 80.9 & 96.1 & 87.0 & 89.9 & 80.3 & 85.9 & \textbf{85.5} \\
    
    \bottomrule
    \end{tabular}}
    \label{tab:results of commonsense compare others}
\end{table*}

\begin{table*}[ht]
    \centering
    \renewcommand\arraystretch{0.95} 
    \caption {Results with DeBERTaV3-base fine-tuned on GLUE development set. ``FT'' represents fully fine-tuning.}
    \resizebox{\textwidth}{!}{
    \begin{tabular}{l | c| c c c c c c c c |>{\columncolor{gray!10}}c}
    \toprule
         \multirow{2}{*}{\textbf{Method}} &  \multirow{2}{*}{\textbf{\% Params}} & \textbf{MNLI} & \textbf{SST-2} &\textbf{CoLA} & \textbf{QQP} & \textbf{QNLI} & \textbf{RTE} & \textbf{MRPC} & \textbf{STS-B} & \textbf{All}\\
         & & Acc & Acc & Mcc & Acc & Acc & Acc & Acc & Corr & Avg. \\ 
         \midrule
        
        FT & 100\% & 89.90 & 95.63 & 69.19 & 91.87 & 94.03 & 83.75 & 90.20 & 91.60 & 88.27 \\ \midrule
        
        (IA)$^3$ & 0.03\% & 89.44 & 95.52 & 67.01 & 89.01 & 91.80 & 79.42 & 88.23 & 90.79 & 86.40 \\  
         
         SSL & 0.02\% & 88.35 & 95.07 & 66.64 & 88.19 & 90.10 & 82.31 & 88.68 & 90.13 & 86.18 \\
        
         SSB & 0.05\% & 89.86 & 95.53 & 67.82 & 89.87 & 93.41 & 83.75 & 88.72 & 90.94 & 87.49 \\
         
        BitFit & 0.05\% & 89.37 & 94.84 & 66.96 & 88.41 & 92.24 & 78.80 & 87.75 & 91.35 & 86.21 \\ \midrule
        
         Series & 0.17\% & 90.10 & 95.41 & 67.65 & 91.19 & 93.52 & 83.39 & 89.25 & 91.31 & 87.73\\
        
        PAdapter & 0.16\% & 89.89 & 94.72 & 69.06 & 91.05 & 93.87 & 84.48 & 89.71 & 91.38 & 88.02\\

         LoRA & 0.18\% & 90.03 & 93.92 & 69.15 & 90.61 & 93.37 &  87.01 & 90.19 & 90.75 & 88.13  \\  

        AdaLoRA & 0.18\% & 90.66 & 95.80 & 70.04 & 91.78 & 94.49 & 87.36 & 90.44 & 91.63 & 88.86 \\
        
        FLoRA & 0.18\% & 90.60 & 96.00 & 70.20 & 91.40 & 94.46 & 88.09 & 90.93 & 91.96 & 89.21 \\
        
        DoRA & 0.22\% & 90.21 & 94.38 & 69.33 & 90.84 & 93.26 & 86.94 & 90.19 & 91.34 & 88.31\\

         \rowcolor{gray!20}

         LoRA-TSD & 0.18\% & 90.26 & 96.10 & 72.41 & 91.65 & 94.77 & 90.03 & 91.91 & 91.89 & \textbf{89.88} \\

        \bottomrule
    \end{tabular}
    }
    \label{tab: deberta results comparison}
\end{table*}

In the main text, we only mainly compared the results of our methods and LoRA because they are designed to further harness the potential of TSDs captured by LoRA. Here, we further compare the performance of LoRA-TSD with other methods.

Specifically, we compare BitFit \cite{zaken2021bitfit}, Prompt Learning (Prefix) \cite{li2021prefix}, Series Adapter (Series) \cite{houlsby2019parameter}, Adapter proposed by \cite{pfeiffer2020adapterfusion} (PAdapter), Parallel Adapter (Parallel) \cite{he2021towards}, (IA)$^3$ \cite{liu2022few}, SSL \cite{si2024see}, SSB \cite{si2024see}, AdaLoRA \cite{zhang2022adaptive}, PISSA \cite{meng2024pissa}, MiLoRA \cite{wang2024milora}, FLoRA \cite{si2024flora}
and DoRA \cite{liu2024dora}. We follow the settings detailed in our paper, and the results are shown in Tables. \ref{tab:results of commonsense compare others}-\ref{tab: deberta results comparison}.

The results clearly show that even compared to state-of-the-art (SOTA) methods like DoRA and FLoRA, LoRA-TSD holds its ground impressively. Additionally, two phenomena are noteworthy:

\begin{itemize}
    \item When fine-tuning LLaMA-7B, at a rank of 4, all methods except LoRA-TSD perform poorly under low parameter budgets, while LoRA-TSD achieves results comparable to those with higher parameter settings. This further highlights the superiority of LoRA-TSD, which stimulates performance on downstream tasks by unleashing the potential of TSDs.
    \item Compared with LoRA derivatives, LoRA-TSD introduces significantly fewer additional parameters, such as nearly 130 times fewer than FLoRA and 700 times fewer than DoRA at rank 8 when fine-tuning LLaMA-7B, yet achieves better or comparable results.
\end{itemize}

Revisiting the objective of Parameter-Efficient Fine-Tuning, the key question is whether it is possible to achieve performance comparable to fully fine-tuning while training with significantly fewer parameters. 
Through extensive experimentation and comparison with other methods, LoRA-TSD maintains excellent performance even with a minimal parameter budget, where most methods would be constrained by the amount of parameters available. 
This demonstrates that LoRA-TSD significantly surpasses other methods in terms of parameter utilization efficiency. Therefore, we conclude that LoRA-TSD exemplifies an ideal realization of PEFT goals.

\vfill

\end{document}